\documentclass[preprint,12pt]{elsarticle}

\usepackage{amsmath}
\usepackage{graphicx}%
\usepackage{subcaption}
\usepackage{multirow}%
\usepackage{amsmath,amssymb,amsfonts}%
\usepackage{mathrsfs}%
\usepackage[title]{appendix}%
\usepackage{xcolor}%
\usepackage{textcomp}%
\usepackage{manyfoot}%
\usepackage{threeparttable}
\usepackage{booktabs}%
\usepackage{algorithm}%
\usepackage{array} 
\usepackage{algorithmicx}%
\usepackage{caption}
\usepackage{algpseudocode}%
\usepackage{listings}%
\usepackage{graphics} 
\usepackage{graphicx}
\usepackage{epsfig} 
\usepackage{mathptmx} 
\usepackage{times} 
\usepackage{amssymb}  
\usepackage{microtype}
\usepackage{algorithm}
\usepackage{float}
\usepackage{makecell}

\journal{Robotics and Autonomous Systems}

\begin{document}

\begin{frontmatter}



\title{Design and Dimensional Optimization of Legged Structures for Construction Robots}


\author[1]{Xiao Liu}
\author[1,2]{Xianlong Yang}
\author[1,3]{Weijun Wang}
\author[1,3]{Wei Feng\corref{cor}}

\cortext[cor]{Corresponding author. Email: wei.feng@siat.ac.cn}

\address[1]{Shenzhen Institute of Advanced Technology, Chinese Academy of Sciences, No. 1068 Xueyuan Avenue, Shenzhen 518055, Guangdong, China}
\address[2]{School of Transportation and Logistics Engineering, Wuhan University of Technology, No. 122, Heping Avenue, Wuhan 430070, Hubei, China}
\address[3]{University of Chinese Academy of Sciences, No. 1 Yanqi Lake East Road, Beijing 101408, China}

\begin{abstract}
Faced with complex and unstructured construction environments, wheeled and tracked robots exhibit significant limitations in terrain adaptability and flexibility, making it difficult to meet the requirements of autonomous operation. Inspired by ants in nature, this paper proposes a leg configuration design and optimization method tailored for construction scenarios, aiming to enhance the autonomous mobility of construction robots. Considering common challenges on construction sites such as uneven terrain, elevation changes, narrow spaces, and dynamic obstacles, this study analyzes the full working conditions of the robotic leg during both swing and stance phases. In response to common challenges in construction sites—such as uneven terrain, height variations, narrow spaces, and dynamic obstacles—this paper analyzes the full operational motion performance of the leg during both swing and stance phases. First, based on kinematic modeling and multi-dimensional workspace analysis, the concept of an "improved workspace" is introduced, and graphical methods are used to optimize the leg dimensions during the swing phase. Furthermore, a new concept of "average manipulability" is introduced based on the velocity Jacobian matrix, and numerical solutions are applied to obtain the leg segment ratio that maximize manipulability. To overcome the difficulties associated with traditional analytical methods, virtual prototype simulations are conducted in ADAMS to explore the relationship between the robot body's optimal flexibility and leg segment proportions. In summary, the leg segment proportions with the best comprehensive motion performance are obtained. This study presents the first multi-dimensional quantitative evaluation framework for leg motion performance tailored for construction environments, providing a structural design foundation for legged construction robots to achieve autonomous mobility in complex terrains.
\end{abstract}

\begin{graphicalabstract}
\centering
\includegraphics[width=0.9\textwidth]{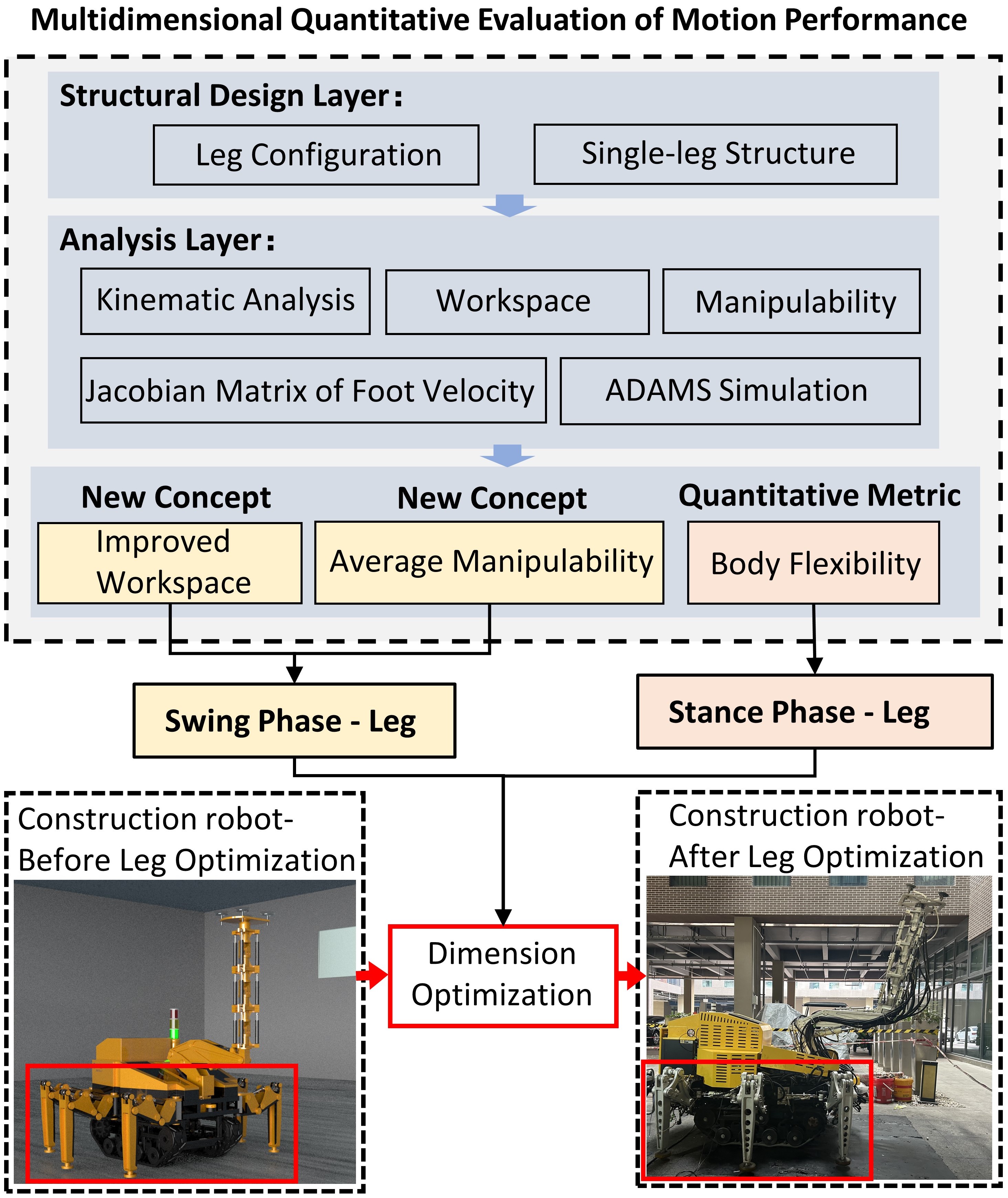}
\end{graphicalabstract}

\begin{highlights}
\item First introduced legged robots into the construction field to meet practical application needs.
\item Proposed an integrated design and optimization method for legs suitable for construction scenarios.
\item Based on walking in a construction environment, we introduced new concepts of "Improved Workspace" and "Average Manipulability".
\item First achieved a multi-dimensional evaluation of leg motion performance in construction environments, with systematic optimization of kinematic performance.
\end{highlights}


\end{frontmatter}



\section{Introduction}

With the rapid development of science and technology, the concept of safe and efficient "Intelligent Construction" has emerged [1-2]. Construction robots, as core equipment in "Intelligent Construction," have received widespread attention. Particularly for construction robots capable of efficient mobility in complex terrains, leg design has become a central research focus. Compared to conventional wheeled or tracked construction robots [3-4], the construction robot studied in this paper utilizes legs alone for movement, offering higher flexibility and terrain adaptability. Therefore, the leg is selected as the research subject. Starting from the requirements of construction robot legs, and drawing inspiration from the leg structures of insects in nature while incorporating advanced multi-legged robot design concepts [5-7], we preliminarily design the overall configuration of the construction robot's legs as well as the structural design of a single leg.

\begin{figure}[h]
   \centering
   \includegraphics[width=0.9\textwidth]{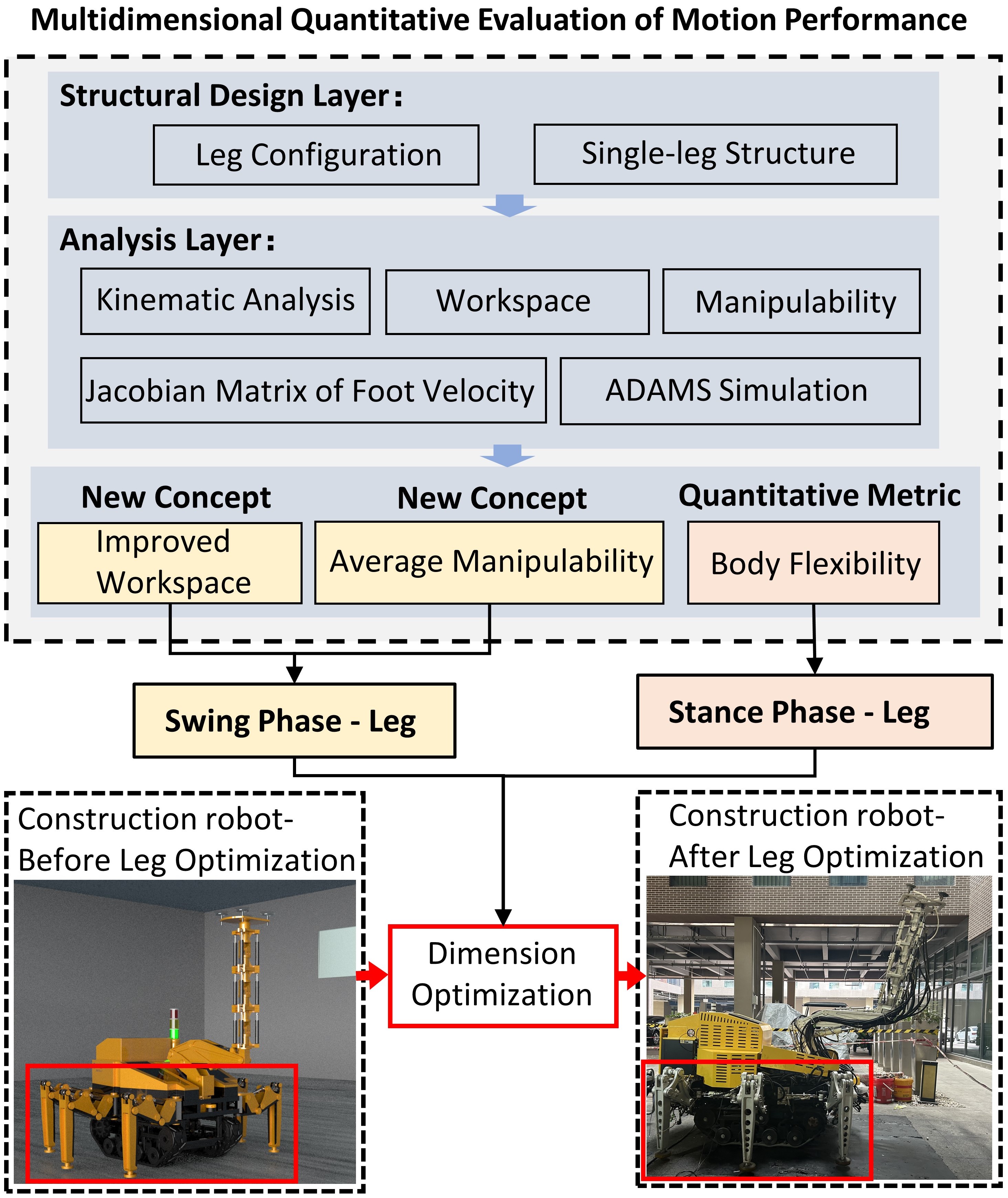}
   \caption{Framework for Leg Design and Dimension Optimization Based on Kinematic Performance}\label{Fig.1}
\end{figure}
   
In the field of structural optimization, extensive research on structural optimization has been conducted by scholars [8-16]. For instance, [17] analyzed the kinematic characteristics of a novel asymmetric parallel biped robot with common hinges and employed particle swarm optimization to optimize the linkage dimensions. [18] develops a virtual equivalent parallel mechanism (VEPM) to model the robot-ground system, optimizing the mechanism’s dimensions based on comprehensive performance analysis to enable a two-wheeled self-balancing robot (TWSBR) to stand, elevate, and overcome obstacles. [19] proposes a structural design and kinematic analysis solution for robotic manipulators in automated 3D concrete construction printing. In [20], a novel method for evaluating robotic dexterity is introduced. [21] presents a design approach for a new 5-DOF hybrid robot for machining, focusing on three performance metrics: workspace-to-machine volume ratio, lateral stiffness, and lower modes. [22] provides kinematic, workspace analysis, and performance evaluation, and furthermore, proposes a dimensional optimization scheme for the manipulator based on motion/force transmission performance and geometric constraints. 

Despite significant advancements in the field of robotic structural optimization, comprehensive analysis and structural optimization studies focusing on legged motion performance in construction scenarios remain limited. This paper aims to address this research gap by introducing legged robots into the construction domain for the first time, driven by the practical requirements of construction robot legs. Focusing on the structural optimization of construction robot legs, this study conducts a multi-level, comprehensive, and systematic quantitative analysis of the kinematic performance of legs under various working conditions. For the first time, it achieves a multi-dimensional quantitative evaluation of leg locomotion performance in construction scenarios, providing theoretical support and technical assurance for the practical application of construction robots.

\section{Leg Structure Design}

In the walking process of a hexapod robot, the state where the leg is in contact with the ground for support is referred to as the stance phase, while the state where the leg is lifted and swings is called the swing phase. The lifted leg forms an open-chain structure relative to the robot's body frame, functioning similarly to a multi-degree-of-freedom serial manipulator. In contrast, the leg in the stance phase, together with the ground and the body frame, constitutes a closed-chain multi-degree-of-freedom parallel mechanism. Therefore, for the swing phase, the focus is on analyzing the influence of leg structural parameters on the kinematic performance of a single leg. For the stance phase, the emphasis is on studying the impact of leg parameters on the overall kinematic performance of the body frame.

To ensure that the construction robot can select footholds within a large target range during walking, the workspace area and manipulability of a single leg are discussed. Additionally, to guarantee that the robot's body frame can achieve an ideal pose during multi-legged support, the influence of leg structures on the spatial flexibility of the body is analyzed. The overall optimization approach is illustrated in Fig. 1, aiming to achieve optimal kinematic performance.

The mobile mechanism of the construction robot in this paper primarily consists of three components: the body frame, legs, and tracked chassis. When the construction robot utilizes multi-legged locomotion, it functions as a multi-legged robot [23]. Based on research, currently developed biomimetic multi-legged robots are mainly categorized into two types: insect-inspired robots (hexapods or more) and mammal-inspired robots (primarily quadrupeds). Mammal-inspired robots exhibit high flexibility, while insect-inspired robots significantly enhance load-bearing capacity and terrain adaptability through multi-point support, making them more suitable for the load requirements of construction scenarios. Inspired by natural insects like ants and tiger beetles, a hexapod insect-inspired configuration is adopted. This design combines a wide range of motion for its legs with enhanced body flexibility. Moreover, its low center of gravity during movement ensures superior stability.

The hind leg of an ant consists of the coxa, femur, tibia, and tarsus [24-25], as shown in Fig.2. Among these, the tarsus serves as an additional structure that primarily enhances crawling stability by increasing the contact area between the leg and the ground. The body, coxa, femur, and tibia are connected through three rotational joints: the root joint, hip joint, and knee joint. Ants primarily achieve overall body movement by actuating the rotation of the root joint, hip joint, and knee joint [26-27].

\begin{figure}[h]
   \centering
   \includegraphics[width=0.9\textwidth]{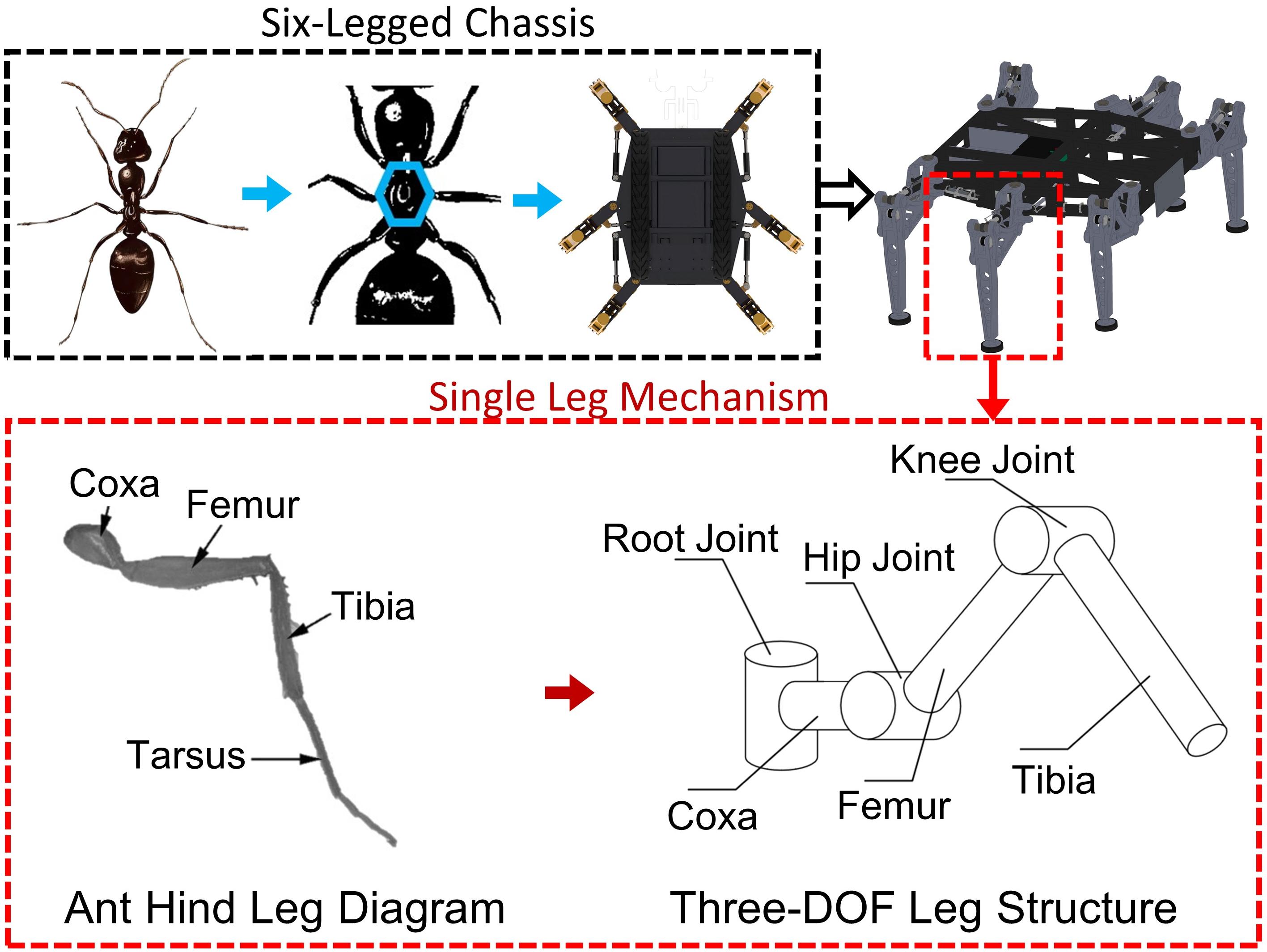}
   \caption{Design Diagram of Leg Structure}\label{Fig.2}
\end{figure}
   
By simulating arthropods such as ants, a biomimetic three-degree-of-freedom leg structure composed of the coxa, femur, and tibia is constructed, as shown in Fig. 2. For clarity, the three rotational joints of the robot leg are referred to as the root joint, hip joint, and knee joint. The leg functions as a three-degree-of-freedom serial manipulator, capable of meeting the ground contact requirements of the foot within a specific range during walking. A hexapod robot in motion can be regarded as a hybrid mechanism consisting of multiple branched parallel structures and multiple serial mechanisms. Assuming ideal conditions where there is no relative motion between the foot and the ground for legs in the stance phase, the contact between the foot and the ground can be treated as a point contact, equivalent to a three-degree-of-freedom spherical joint, while each of the three joints in a single leg is a one-degree-of-freedom revolute joint. If the number of legs in the stance phase is denoted as N, the total degrees of freedom (F) of the robot's mobile mechanism as a whole is determined accordingly.

\begin{equation}
\label{(1)}
F=\sum_{i=1}^pf_i-\sum_{i=1}^L\lambda_i-f_p-F^{\prime}+\lambda_0
\end{equation}
Among these:
- \( p \) is the number of kinematic pairs, \( p = 4N \);
- \( f_i \) is the degree of freedom of the \( i \)-th kinematic pair;
- \( L \) is the number of independent closed loops in the parallel mechanism, \( L = N-1 \);
- \( \lambda_i \) is the number of constraint conditions for the \( i \)-th independent closed loop, \( \lambda_i = 6 \);
- \( f_p \) and \( F' \) are the redundant degrees of freedom and the independent degrees of freedom, both equal to 0;
- \( \lambda_0 \) is the number of redundant constraints, \( \lambda_0 = 0 \).

Simplifying Equation (1) yields:
\begin{equation}
\label{(2)}
F=3N+3N-(N-1)\times6=6
\end{equation}

It can be concluded that the total degrees of freedom \( F \) of the mobile mechanism is always 6, indicating that the robot functions as a multi-loop parallel mechanism with a constant six degrees of freedom during walking, regardless of the number of legs in the stance phase. The robot's body can achieve the desired pose within a specific range, demonstrating that the leg design preliminarily meets the requirements.

\section{Kinematic Analysis Of The Leg}
\subsection{Swing Phase of the Leg}
During the walking process of a hexapod robot, the swing phase leg mechanism can be considered a three-degree-of-freedom serial manipulator. Let \( \theta_1 \), \( \theta_2 \), and \( \theta_3 \) represent the rotational angles of the root joint, hip joint, and knee joint, respectively, and \( l_1 \), \( l_2 \), and \( l_3 \) represent the lengths of the coxa, femur, and tibia, respectively. The D-H coordinate frames are established for each leg segment. The simplified model of the swing phase leg mechanism and the D-H coordinate frames are shown in Fig.3.
\begin{figure}[htb]
      \centering
      \includegraphics[width=0.7\columnwidth]{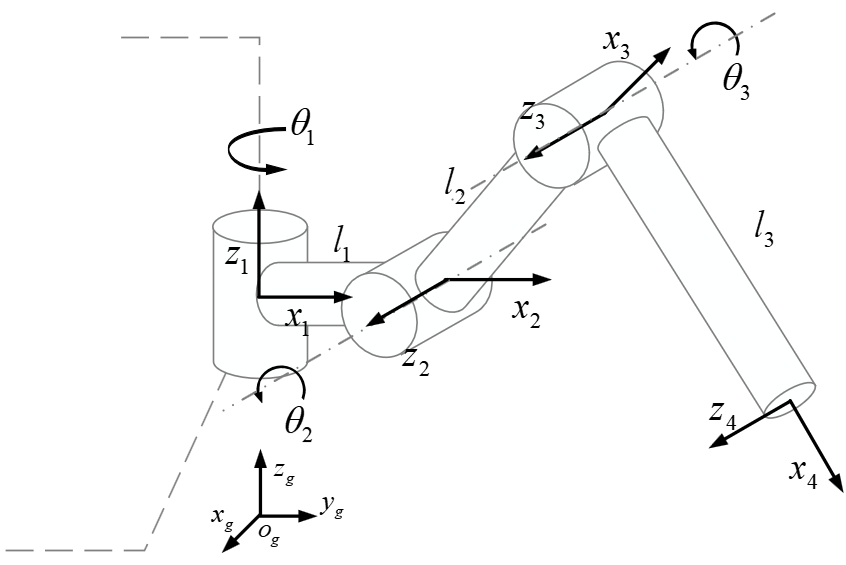}
      \caption{Simplified Model of the Leg in Swing Phase}
      \label{Fig.3}
   \end{figure}
   
Among these:
- \( O_i \) is the origin of each coordinate system;
- \( z_i \) is the axis of joint \( i \);
- \( x_i \) is the common perpendicular between the axis of joint \( i-1 \) and the axis of joint \( i \).
Based on the simplified diagram of the mechanism and the D-H coordinate frames, the D-H parameter table can be listed as shown in Table 1.
\begin{table}[ht]
    \centering
    \caption{D-H Parameters of the Leg in Swing Phase}
    \label{Table 1}
    \begin{tabular}{c c c c c}
        \hline
        Link $i$ & Joint Angle $\theta_i$ & Twist Angle $\alpha_i$ & Offset $d_i$ & Length $a_i$ \\
        \hline
        1 & $\theta_1$ & $\frac{\pi}{2}$ & 0 & $l_1$ \\
        2 & $\theta_2$ & 0 & 0 & $l_2$ \\
        3 & $\theta_3$ & 0 & 0 & $l_3$ \\
        \hline
    \end{tabular}
\end{table}

Among these:
- \( \alpha_i \), \( a_i \), and \( d_i \) are determined by the leg structure and remain constant.
- The forward kinematic analysis of the swing phase leg involves solving for the position of the foot end given the rotational angles of each joint.
- Substituting the aforementioned D-H parameters into the homogeneous transformation matrices between adjacent joint coordinate systems yields:
\begin{equation}
\label{(3)}
T_1^0=
\begin{bmatrix}
\cos\theta_1 & 0 & \sin\theta_1 & l_1\cos\theta_1 \\
\sin\theta_1 & 0 & -\cos\theta_1 & l_1\sin\theta_1 \\
0 & 1 & 0 & 0 \\
0 & 0 & 0 & 1
\end{bmatrix}
\end{equation}
\begin{equation}
\label{(4)}
T_2^1=
\begin{bmatrix}
\cos\theta_2 & -\sin\theta_2 & 0 & l_2\cos\theta_2 \\
\sin\theta_2 & \cos\theta_2 & 0 & l_2\sin\theta_2 \\
0 & 0 & 1 & 0 \\
0 & 0 & 0 & 1
\end{bmatrix}
\end{equation}
\begin{equation}
\label{(5)}
T_3^2=
\begin{bmatrix}
\cos\theta_3 & -\sin\theta_3 & 0 & l_3\cos\theta_3 \\
\sin\theta_3 & \cos\theta_3 & 0 & l_3\sin\theta_3 \\
0 & 0 & 1 & 0 \\
0 & 0 & 0 & 1
\end{bmatrix}
\end{equation}
During the forward kinematic analysis, let the coordinates at the root joint of the leg serve as the base coordinates, and the position coordinates of the foot end be represented by the vector \(\begin{bmatrix} p_x & p_y & p_z \end{bmatrix}^T\). The transformation matrix for the fixed coordinate system attached to the foot end is given by:
\begin{equation}
\label{(6)}
T_0^3=T_0^1T_1^2T_3^2=
\begin{bmatrix}
n_x & o_x & a_x & p_x \\
n_y & o_y & a_y & p_y \\
n_z & o_y & a_z & p_z \\
0 & 0 & 0 & 1
\end{bmatrix}
\end{equation}
Substituting equations (3), (4), and (5) yields the relationship between the foot-end coordinates and the joint angles.
\begin{equation}
\label{(7)}
\begin{cases}
p_x=\cos\theta_1\left(l_3\cos\left(\theta_2+\theta_3\right)+l_2\cos\theta_2+l_1\right) \\
p_y=\sin\theta_1\left(l_3\cos\left(\theta_2+\theta_3\right)+l_2\cos\theta_2+l_1\right) \\
p_z=l_3\sin\left(\theta_2+\theta_3\right)+l_2\sin\theta_2 & 
\end{cases}
\end{equation}
The above represents the forward kinematic solution for the swing phase leg.

The inverse kinematic analysis of the swing phase leg in the robot involves solving for the rotational angles of each joint given the position variables of the foot end. Using the forward kinematic solution from equation (7), the inverse kinematic analysis can be performed through operations between transformation matrices. Since the inverse kinematic equations may have multiple solutions, and considering the actual working state of the leg, the joint angle ranges are provisionally set as follows: \( \theta_1 \in [-\pi/2, \pi/2] \), \( \theta_2 \in [-\pi/2, \pi/2] \), \( \theta_3 \in [-\pi, 0] \). Utilizing \( (T_1^0)^{-1} T_3^0 = T_1^1 T_3^1 \) yields:
\begin{equation}
\label{(8)}
\begin{cases}
 & \theta_1=\arctan\left(\frac{p_y}{p_x}\right) \\
 & \theta_2=\arcsin\left(\frac{p_z}{\sqrt{\tau+l_1^2}}\right)+\arccos\left(\frac{\tau+l_1^2+l_2^2-l_3^2}{2l_2\sqrt{\tau+l_1^2}}\right) \\
 & \theta_3=-\arccos\left(\frac{\tau+l_1^2-l_2^2-l_3^2}{2l_2l_3}\right)
\end{cases}
\end{equation}
Among these:
\begin{equation}
\label{(9)}
\tau=p_x^2+p_y^2+p_z^2-2l_1\sqrt{p_x^2+p_y^2}
\end{equation}
The above represents a set of inverse kinematic solutions for the swing phase leg.
\subsection{Stance Phase of the Leg}
During the walking process of the construction robot, the multiple legs in the stance phase, together with the robot's body frame and the ground, form a parallel mechanism. At this point, the pose of the robot's body frame is related to the pose of the legs in the stance phase [28]. Therefore, the kinematic analysis of the stance phase legs involves solving the relationship between the pose of the robot's body frame and the joint angles of the legs in the stance phase, which belongs to the kinematic analysis problem of parallel multi-loop mechanisms. A kinematic modeling and analysis of the stance phase legs is now conducted.

A fixed coordinate system is established at the center of the robot's body frame, while the other coordinate systems can be referenced from the kinematic model of the swing phase legs. The simplified model of the stance phase leg mechanism is shown in Fig.4.
\begin{figure}[htb]
      \centering
      \includegraphics[width=0.7\columnwidth]{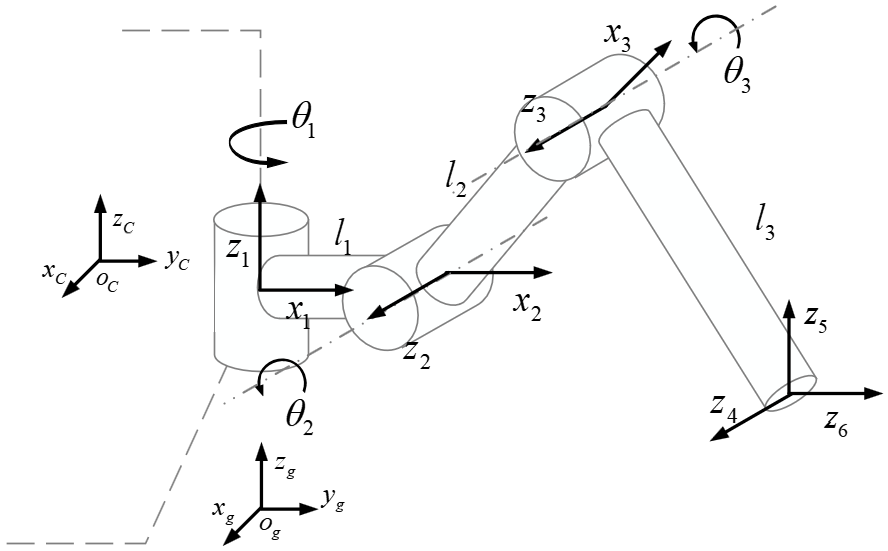}
      \caption{Simplified Model of the Leg in Stance Phase}
      \label{Fig.4}
   \end{figure}
The pose transformation matrix of the robot's body frame center relative to the global coordinate system is:
\begin{equation}
    \resizebox{0.8\hsize}{!}{$
    \label{(10)}
    \begin{split}
    &  T_P= \\  
    &\begin{bmatrix}
    \cos\alpha\cos\beta & \cos\alpha\sin\beta\cos\gamma-\sin\alpha\cos\gamma & \cos\alpha\sin\beta\cos\gamma+\sin\alpha\sin\gamma & X_C \\
    \sin\alpha\cos\beta & \sin\alpha\sin\beta\cos\gamma+\cos\alpha\cos\gamma & \sin\alpha\sin\beta\cos\gamma-\cos\alpha\sin\gamma & Y_C \\
    -\sin\beta & \cos\beta\sin\gamma & \cos\beta\cos\gamma & Z_C \\
    0 & 0 & 0 & 1
  \end{bmatrix}
  \end{split}
        $}
\end{equation}

Among these:
- \( X_C \), \( Y_C \), and \( Z_C \) represent the coordinates of the robot's body frame center in the global coordinate system.
- \( \alpha \), \( \beta \), and \( \gamma \) represent the angles of rotation of the body frame center about the \( X \), \( Y \), and \( Z \) axes of the global coordinate system, respectively.

The tripod gait, also known as the triangular gait, refers to a walking pattern where a hexapod robot has three legs in the swing phase and three legs in the stance phase during locomotion. This gait balances stability and speed. The construction robot discussed in this paper has specific requirements for both speed and stability, so it employs the tripod gait as the basis for its walking pattern. For the kinematic analysis of the legs in the stance phase, the three legs in the stance phase are designated as legs 1, 2, and 3.

Combining the kinematic analysis of the swing phase legs, let the coordinates of the foot end be \(\begin{bmatrix} p_x & p_y & p_z \end{bmatrix}^T\). Then:
\begin{equation}
\label{(11)}
\begin{bmatrix}
p_x & p_y & p_z & 1
\end{bmatrix}^T=T_CT_0^CT_1^0T_2^1T_3^2
\begin{bmatrix}
0 & 0 & 0 & 1
\end{bmatrix}^T
\end{equation}

In the equation, \( T_0^C \) represents the transformation matrix between the root joint coordinate system and the center coordinate system, which is determined by the dimensional parameters of the robot. Given the previous information, \( T_0^1 \), \( T_1^2 \), and \( T_2^3 \) are known. Therefore, \( T_0^C T_1^0 T_2^1 T_3^2 \begin{bmatrix} 0 & 0 & 0 & 1 \end{bmatrix}^T \) can be expressed as \(\begin{bmatrix} u & v & w & 1 \end{bmatrix}^T\). Thus, Equation (11) can be represented as:
\begin{equation}
\label{(12)}
\begin{bmatrix}
p_x & p_y & p_z & 1
\end{bmatrix}^T=T_C
\begin{bmatrix}
u & v & w & 1
\end{bmatrix}^T
\end{equation}

For the three legs (1, 2, and 3) in the stance phase, the relationship can be derived from Equation (12):
\begin{equation}
\label{(13)}
\begin{bmatrix}
p_{x1} & p_{x2} & p_{x3} \\
p_{y1} & p_{y2} & p_{y3} \\
p_{z1} & p_{z2} & p_{z3} \\
1 & 1 & 1
\end{bmatrix}=T_C
\begin{bmatrix}
u_1 & u_2 & u_3 \\
v_1 & v_2 & v_3 \\
w_1 & w_2 & w_3 \\
1 & 1 & 1
\end{bmatrix}
\end{equation}

For Equation (13), it can be seen that when the foot-end coordinates and joint angles of the legs in the stance phase are known, \( T_C \) can be determined, which represents the pose parameters of the robot's body frame center. This is the forward kinematic solution for the stance phase legs. Conversely, when the pose parameters of the robot's body frame center and the foot-end coordinates of the stance phase legs are known, the joint angles of each leg in the stance phase can be determined. This is the inverse kinematic solution for the stance phase legs.

\section{Dimension Optimization Based On The Improved Workspace}
\subsection{Workspace of the Leg in Swing Phase}
The workspace reflects the range of motion of the leg's foot-end and serves as a key indicator for evaluating the leg's mobility [29]. It also provides a basis for assessing the rationality of the leg's structural design. Clarifying this concept lays the foundation for subsequent optimization research on the leg. Considering the actual working conditions of the robotic leg in construction, the ranges of motion for the root joint, hip joint, and knee joint are defined as shown in Table 2.

\begin{table}[ht]
    \centering
    \caption{Range of Joint Angle Values}
    \label{Table 2}
    \begin{tabular}{lccc}
        \toprule
        Joint Angle & $\theta_1$ & $\theta_2$ & $\theta_3$ \\
        \midrule
        Minimum Value (rad) & $-\pi/4$ & $-\pi/3$ & $-3\pi/4$ \\
        Maximum Value (rad) & $\pi/4$ & $\pi/3$ & $0$ \\
        \bottomrule
    \end{tabular}
\end{table}

Based on the range of motion for each joint angle, the Monte Carlo method is used to solve for the workspace of the leg. Assuming the lengths of the leg segments are as follows: coxa \( l_1 = 200 \, \text{mm} \), femur \( l_2 = 400 \, \text{mm} \), and tibia \( l_3 = 400 \, \text{mm} \). The workspace of the swing phase leg is illustrated in Fig.5 using MATLAB.
\begin{figure}[htb]
      \centering
      \includegraphics[width=0.6\columnwidth]{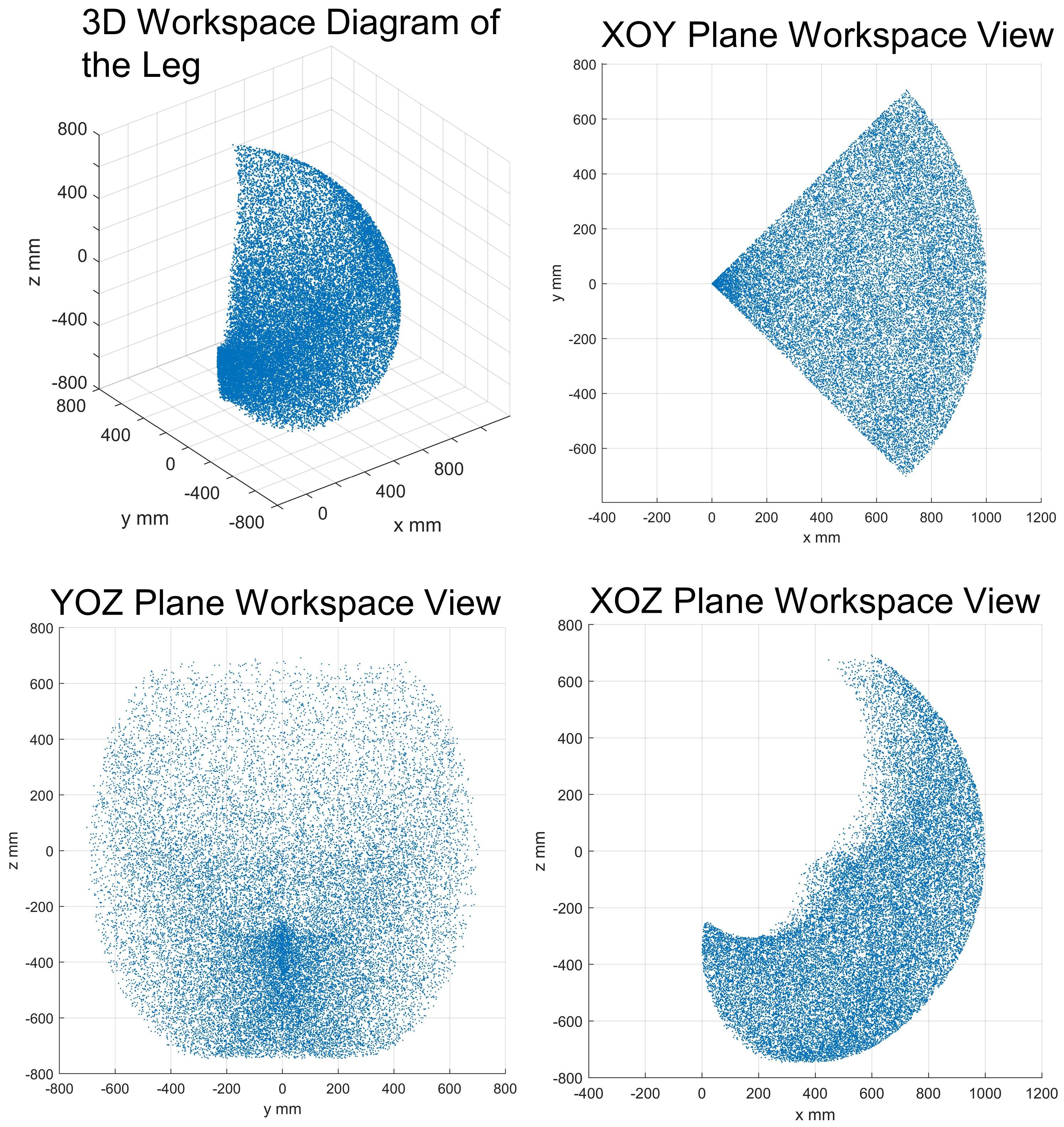}
      \caption{Point Cloud Diagram of the Leg Workspace in Swing Phase}
      \label{Fig.5}
   \end{figure}
   
The workspace contour generated by the Monte Carlo method is clear, providing an intuitive representation of the swing phase leg's foot-end operational area. This serves as a reference for subsequent structural optimization.
\subsection{Foot Workspace in the 2D Plane}
Combining the point cloud diagram of the workspace obtained using the Monte Carlo method with the kinematic model of the swing phase leg, it can be observed that when the total length of the leg is constant, the size of the foot-end workspace projection on the XOY plane is primarily determined by the rotation angle \( \theta_1 \) of the root joint. Additionally, for any given value of \( \theta_1 \) within its defined range, the point cloud diagram of the two-dimensional workspace formed by the hip and knee joints remains the same. Therefore, by restricting the angle \( \theta_1 \), the two-dimensional workspace of the foot end in the plane can be used as the primary target for optimization. Setting \( \theta_1 = 0 \), and based on the forward kinematic solution of the swing phase leg, the coordinate equation of the foot-end point in the two-dimensional plane coordinate system can be derived as:
\begin{equation}
\label{(14)}
p\left(x,z\right)=
\begin{bmatrix}
l_1+l_2\cos\theta_2+l_3\cos\left(\theta_2+\theta_3\right) \\
l_2\sin\theta_2+l_3\sin\left(\theta_2+\theta_3\right)
\end{bmatrix}
\end{equation}
The ranges of the rotation angles \( \theta_2 \) and \( \theta_3 \) for the hip joint and knee joint are shown in Table 3.
\begin{table}[ht]
    \centering
    \caption{Value Ranges of Joints 2 and 3}
    \label{Table 3}
    \resizebox{7cm}{!}
    {%
        \begin{tabular}{lcc}
           \hline
             Joint Angle & $\theta_2$ & $\theta_3$ \\
           \hline
             Minimum Value (rad) & $-\pi/3$ & $-3\pi/4$ \\
           \hline
             Maximum Value (rad) & $\pi/3$ & $0$ \\
           \hline
        \end{tabular}%
    }
\end{table}
Assuming the initial lengths of the leg segments before optimization are as follows: coxa \( l_1 = 200 \, \text{mm} \), femur \( l_2 = 400 \, \text{mm} \), and tibia \( l_3 = 400 \, \text{mm} \). Based on the coordinate parameter equations of the foot end, the two-dimensional workspace envelope of the foot end is illustrated in Fig.6 using MATLAB.
\begin{figure}[htb]
      \centering
      \includegraphics[width=0.5\columnwidth]{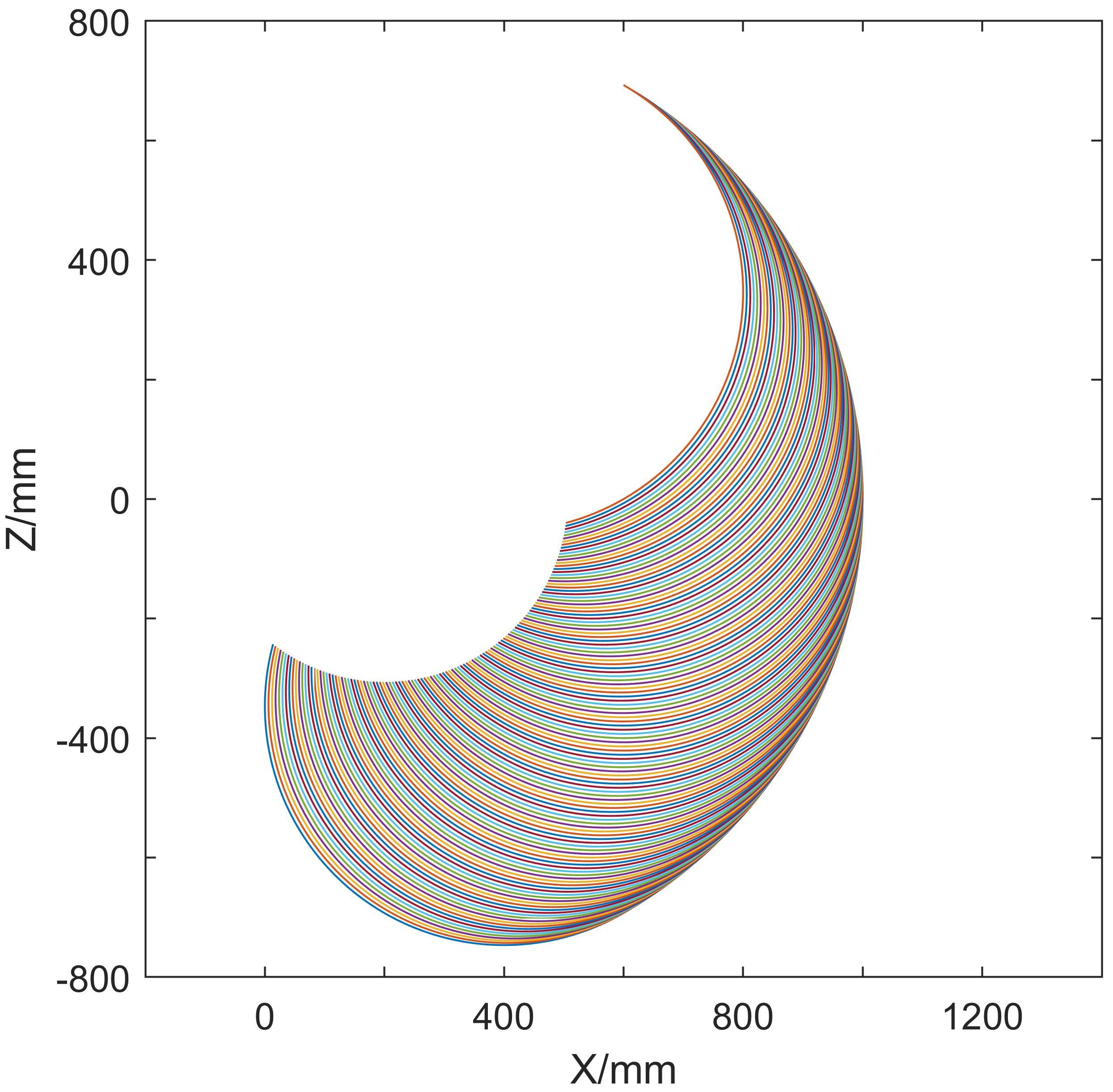}
      \caption{Schematic Diagram of the Foot Workspace in the 2D Plane}
      \label{Fig.6}
   \end{figure}
\subsection{Improved Foot-End Workspace}
However, in the actual working process of the swing phase leg, the motion of the leg can interfere with the body, meaning that the knee joint angle \( \theta_3 \) cannot achieve its ideal range. Therefore, the reachable workspace of the foot end is only a portion of the theoretical value. An improved foot-end workspace is proposed by adjusting the range of \( \theta_3 \) and correlating its limit values with \( \theta_2 \). This ensures that during the working process of the swing phase leg, the angle between the tibia and the horizontal plane is always greater than or equal to 90° (the positive direction is determined by the right-hand screw rule, consistent with the positive direction of the joint angle \( \theta \)).

When the swing phase leg operates within the improved workspace, interference between the leg and the body is avoided. Additionally, it ensures that the angle between the tibia and the contact surface remains greater than or equal to 90° when the foot is placed on obstacles of arbitrary height within the designated range. This can enhance the stability of the robot's body during walking to a certain extent. The revised joint rotation angle ranges are shown in Table 4.
\begin{table}[ht]
    \centering
    \caption{Improved Range of Joint Angles}
    \label{Table 4}
    \resizebox{7cm}{!}{%
        \begin{tabular}{lcc}
           \hline
             Joint Angle & $\theta_2$ & $\theta_3$ \\
           \hline
             Minimum Value (rad) & $-\pi/3$ & $-\theta_2 - \pi/2$ \\
           \hline
             Maximum Value (rad) & $\pi/3$ & $0$ \\
           \hline
        \end{tabular}%
     }
\end{table}
With \( l_1 \), \( l_2 \), and \( l_3 \) held constant and the revised joint angle ranges applied, the improved foot-end workspace is presented in Fig.7.
\begin{figure}[htb]
      \centering
      \includegraphics[width=0.5\columnwidth]{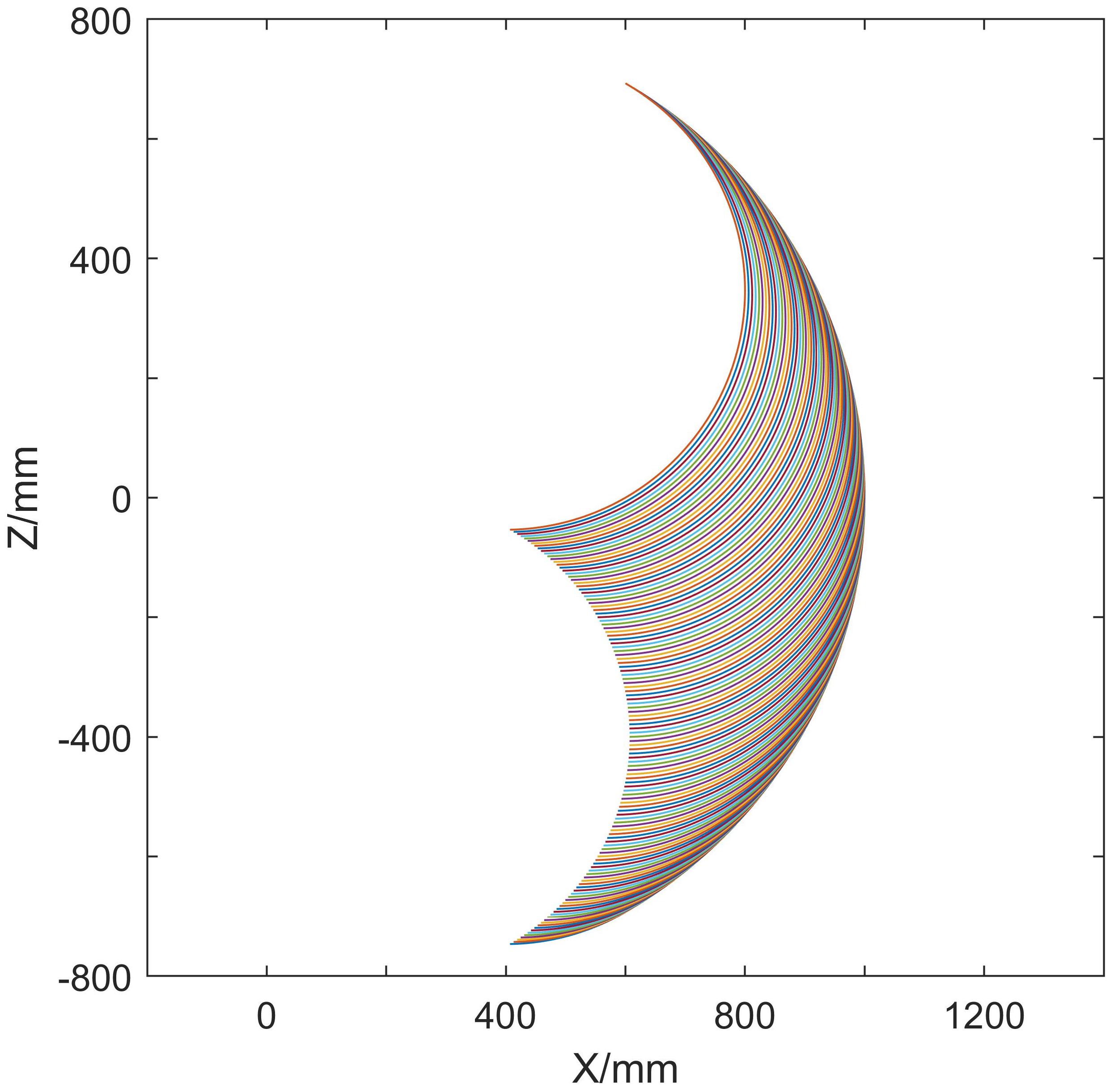}
      \caption{Schematic Diagram of the Improved Foot Workspace}
      \label{Fig.7}
   \end{figure}
It can be seen that by correlating the knee joint angle \( \theta_3 \) with the hip joint angle \( \theta_2 \), the improved workspace more closely approximates the reachable area of the swing phase leg in actual operation. Compared to the original workspace, this improved workspace is more suitable as an indicator for leg dimension optimization.
\subsection{Optimization Results of Workspace Area}
Assuming the total length of the swing phase leg, consisting of the coxa, femur, and tibia, is a fixed value of 1000 mm, i.e., \( l = l_1 + l_2 + l_3 = 1000 \, \text{mm} \). When the total length of the leg remains constant, the leg segment proportions are optimized based on the improved workspace area.

Due to the restriction on the rotation of the root joint \( \theta_1 \), the position of the coxa \( l_1 \) is fixed. Therefore, the relationship between the proportion of \( l_1 \) and the workspace area is first investigated. Keeping the ratio of \( l_2 \) to \( l_3 \) at 1:1, different proportions of \( l_1 \) relative to \( l \) are considered, and the corresponding workspaces are plotted. The values for each leg segment length are shown in Table 5.
\begin{table}[ht]
    \centering
    \caption{Length Values of Each Leg Segment}
    \label{Table 5}
    \resizebox{8cm}{!}{%
        \begin{tabular}{lccc}
           \hline
             $l_1$ Ratio & $l_1$ Length (mm) & $l_2$ Length (mm) & $l_3$ Length (mm) \\
           \hline
             0.05 & 50 & 475 & 475 \\
           \hline
             0.10 & 100 & 450 & 450 \\
           \hline
             0.15 & 150 & 425 & 425 \\
           \hline
             0.20 & 200 & 400 & 400 \\
           \hline
        \end{tabular}%
      }
\end{table}
Based on the above dimensions, schematic diagrams of each workspace are illustrated in Fig.8.
\begin{figure}[htb]
      \centering
      \includegraphics[width=0.5\columnwidth]{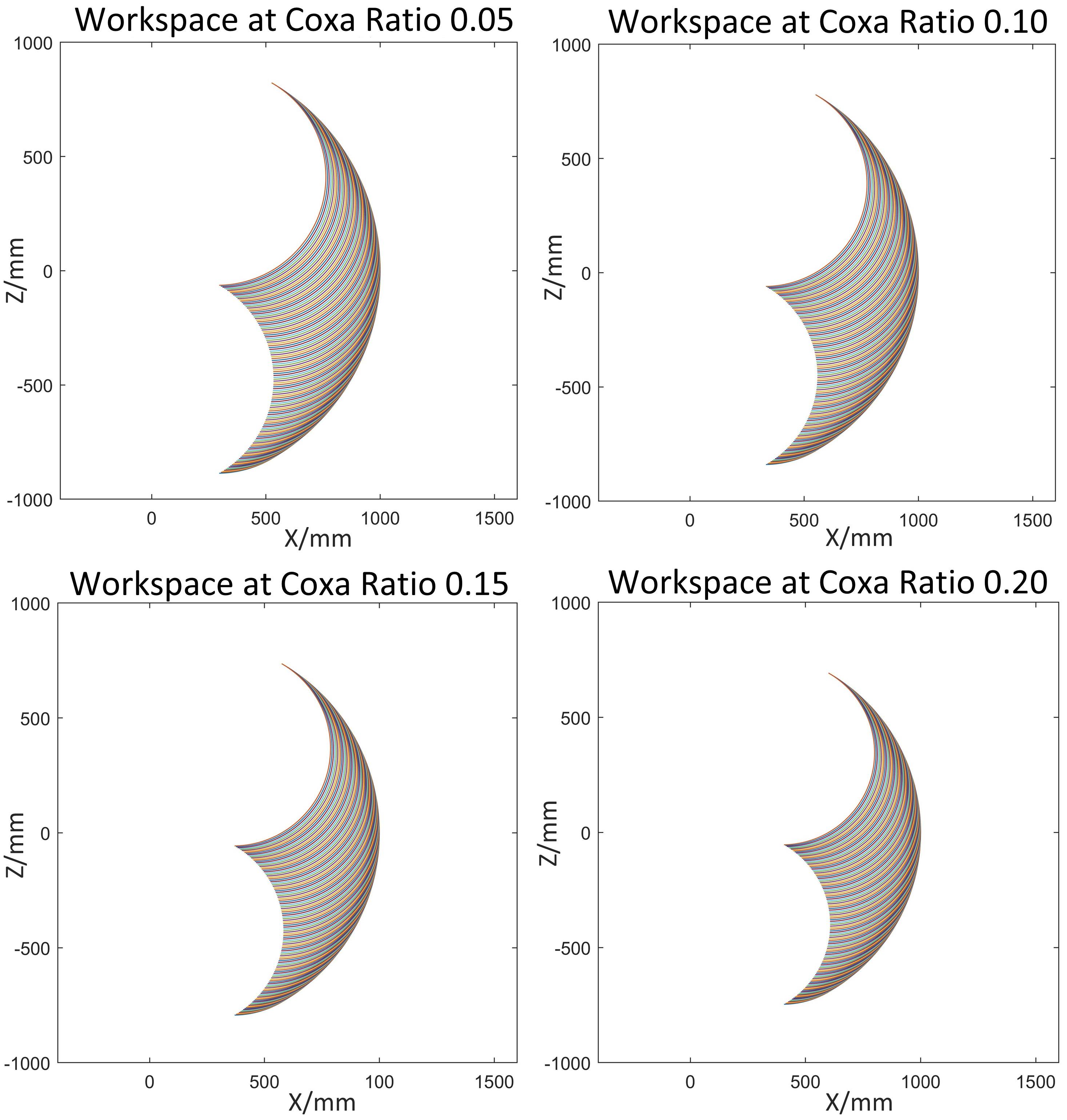}
      \caption{Foot Workspace Diagram with Different Coxa Ratios}
      \label{Fig.8}
   \end{figure}
It can be visually observed from the figure that when the ratio of the femur \( l_2 \) to the tibia \( l_3 \) remains constant, the area of the improved workspace decreases as the ratio of the coxa length \( l_1 \) to the total leg length \( l \) increases.

Next, keeping the ratio of the coxa \( l_1 \) to the total length \( l \) constant, the relationship between the workspace area and the ratios of the femur \( l_2 \) and tibia \( l_3 \) is investigated. Due to the restriction on the coxa rotation, the degrees of freedom of the leg are reduced to 2. With fewer degrees of freedom, the workspace area can be solved using a graphical method. Through geometric analysis, the schematic diagram of the leg's workspace in the two-dimensional plane using the graphical method is shown in Fig.9.
\begin{figure}[htb]
      \centering
      \includegraphics[width=0.35\columnwidth]{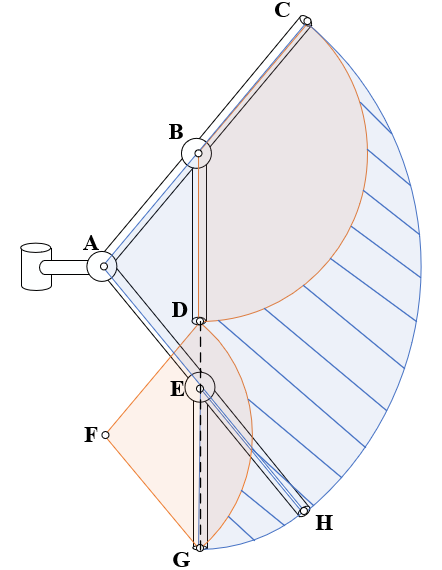}
      \caption{Schematic Diagram of Workspace Area by Graphical Method}
      \label{Fig.9}
   \end{figure}
The required workspace is the shaded area in the figure. Based on the geometric relationship, the area \( S_W \) is given by:
\begin{equation}
\label{(15)}
\begin{split}
S_W=&S_{fanACH}+S_{fanEHG}-S_{fanBCD}-\left(S_{fanFDG}-S_{triFDG}\right)\\
&-S_{triABE}
\end{split}
\end{equation}
Substituting \( l_2 \) and \( l_3 \) yields:
\begin{equation}
\label{(16)}
\begin{split}
S_W=&\frac{(l_2+l_3)^2\pi}{3}+\frac{l_3^2\pi}{12}-[\frac{5l_2^2\pi}{12}+(\frac{l_2^2\pi}{3}-\frac{l_3^2\sqrt{3}}{4})+\frac{l_2^2\sqrt{3}}{4}] \\
=&\frac{2l_2l_3\pi}{3}
\end{split}
\end{equation}
Assuming \( l_2 + l_3 = l' \), when the coxa length \( l_1 \) is determined, \( l' \) is a constant. Substituting into the equation, it can be derived that when \( S_W \) reaches its maximum value, \( l_3 = l_2 = l' / 2 \).

From the above calculations, it is known that when the total length of the swing phase leg is constant, the lower the ratio of the coxa length \( l_1 \) to the total length, the larger the improved workspace area \( S_W \). When the ratio of the coxa is fixed, the improved workspace area \( S_W \) reaches its maximum value when the femur length \( l_2 \) and tibia length \( l_3 \) are equal, i.e., \( l_3 = l_2 = l' / 2 \). Therefore, in designing the legs of a construction robot, the ratio of the coxa length \( l_1 \) can be reduced, and the lengths of the femur and tibia can be made similar. In other words, when the ratio of the coxa length \( l_1 \) is between 0.1 and 0.2, the ratio of the tibia length \( l_3 \) should be between 0.4 and 0.45 to achieve the maximum reachable area for the foot end of the robot's leg.
\section{Dimension Optimization Based on Average Manipulability}
The higher the manipulability value, the greater the robot's flexibility within the workspace, which can partially demonstrate the rationality of the robot's design. Therefore, the manipulability value is used as another quantitative metric for leg dimension optimization. Based on practical considerations, we propose the concept of average manipulability in conjunction with the leg's workspace.
\subsection{Foot Velocity Jacobian Matrix}
For the position analysis of the swing phase leg, the relationship between the end-effector space \( \boldsymbol{p} \) and the joint space \( \boldsymbol{q} \) is defined as:
\begin{equation}
\label{(17)}
p=p(q)
\end{equation}
Differentiating both sides of the equation with respect to time yields:
\begin{equation}
\label{(18)}
\dot{p}=J(q)\dot{q}
\end{equation}
Here, \( \dot{\boldsymbol{p}} \) represents the spatial velocity of the end-effector, which includes both angular and linear velocities. \( \dot{\boldsymbol{q}} \) denotes the joint angular velocity, and \( J(\boldsymbol{q}) \) is the required Jacobian matrix.
\begin{equation}
\label{(19)}
J_{ij}
\begin{pmatrix}
q
\end{pmatrix}=\frac{\partial p_i
\begin{pmatrix}
q
\end{pmatrix}}{\partial q_j},i=1,2,...,6;j=1,2,...,n
\end{equation}
For the swing phase leg, which has three joints, the Jacobian matrix derived from the above equation should theoretically be a \( 6 \times 3 \) matrix. However, since only the linear transformation between the foot-end linear velocity and the joint angular velocities of the leg is required, the resulting Jacobian matrix is a \( 3 \times 3 \) matrix, expressed as follows:
\begin{equation}
\label{(20)}
J
\begin{pmatrix}
q
\end{pmatrix}=
\begin{bmatrix}
\frac{\partial p_x}{\partial\theta_1} & \frac{\partial p_x}{\partial\theta_2} & \frac{\partial p_x}{\partial\theta_3} \\
\frac{\partial p_y}{\partial\theta_1} & \frac{\partial p_y}{\partial\theta_2} & \frac{\partial p_y}{\partial\theta_3} \\
\frac{\partial p_z}{\partial\theta_1} & \frac{\partial p_z}{\partial\theta_2} & \frac{\partial p_z}{\partial\theta_3}
\end{bmatrix}
\end{equation}
Based on the forward kinematic solution of the swing phase leg, substituting Equation (7) into the above equation yields:
\begin{equation}
\label{(21)}
J
\begin{pmatrix}
q
\end{pmatrix}=
\begin{bmatrix}
J_{11}
\begin{pmatrix}
q
\end{pmatrix} & J_{12}
\begin{pmatrix}
q
\end{pmatrix} & J_{13}
\begin{pmatrix}
q
\end{pmatrix} \\
J_{21}
\begin{pmatrix}
q
\end{pmatrix} & J_{22}
\begin{pmatrix}
q
\end{pmatrix} & J_{23}
\begin{pmatrix}
q
\end{pmatrix} \\
J_{31}
\begin{pmatrix}
q
\end{pmatrix} & J_{32}
\begin{pmatrix}
q
\end{pmatrix} & J_{33}
\begin{pmatrix}
q
\end{pmatrix}
\end{bmatrix}
\end{equation}
Among them:
\begin{equation}
\label{(22)}
\begin{cases}
 & J_{11}\left(q\right)=-\sin\theta_1\left(l_3\cos\left(\theta_2+\theta_3\right)+l_2\cos\theta_2+l_1\right) \\
 & J_{12}\left(q\right)=\cos\theta_1\left(-l_3\sin\left(\theta_2+\theta_3\right)-l_2\sin\theta_2\right) \\
 & J_{13}\left(q\right)=\cos\theta_1\left(-l_3\sin\left(\theta_2+\theta_3\right)\right) \\
 & J_{21}\left(q\right)=\cos\theta_1\left(l_3\cos\left(\theta_2+\theta_3\right)+l_2\cos\theta_2+l_1\right) \\
 & J_{22}\left(q\right)=\sin\theta_1\left(-l_3\sin\left(\theta_2+\theta_3\right)-l_2\sin\theta_2\right) \\
 & J_{23}\left(q\right)=\sin\theta_1\left(-l_3\sin\left(\theta_2+\theta_3\right)\right) \\
 & J_{31}\left(q\right)=0 \\
 & J_{32}\left(q\right)=l_3\cos\left(\theta_2+\theta_3\right)+l_2\cos\theta_2 \\
 & J_{33}\left(q\right)=l_3\cos\left(\theta_2+\theta_3\right)
\end{cases}
\end{equation}
The above equation represents the velocity Jacobian matrix for the foot-end of the swing phase leg.
\subsection{Manipulability of the Robot}
According to the definition of manipulability, for a non-redundant robotic arm, its manipulability is the absolute value of the determinant of the velocity Jacobian matrix, expressed as:
\begin{equation}
\label{(23)}
w=\left|\det\left(J\left(q\right)\right)\right|
\end{equation}
Substituting the foot-end velocity Jacobian matrix from Equation (22) into the above equation yields:
\begin{equation}
\label{(24)}
\begin{aligned}
 &\mathrm{W} =|l_2l_3^2\cos^2(\theta_2+\theta_3)\cos^2\theta_1\sin\theta_2\\
 & -l_2^2l_3\sin(\theta_2+\theta_3)\cos^2\theta_1\cos^2\theta_2\\
 & +l_2l_3^2\cos(\theta_2+\theta_3)\sin(\theta_2+\theta_3)\cos^2\theta_1\cos\theta_2\\
 & -l_2l_3^2\cos(\theta_2+\theta_3)\sin(\theta_2+\theta_3)\sin^2\theta_1\cos\theta_2 \\
 & +l_2^2l_3\cos(\theta_2+\theta_3)\cos^2\theta_1\cos\theta_2\sin\theta_2 \\
 & +l_2^2l_3\cos(\theta_2+\theta_3)\sin^2\theta_1\cos\theta_2\sin\theta_2 \\
 & +l_1l_2l_3\cos(\theta_2+\theta_3)\cos^2\theta_1\sin\theta_2 \\
 & -l_1l_2l_3\sin(\theta_2+\theta_3)\cos^2\theta_1\cos\theta_2 \\
 & +l_1l_2l_3\cos(\theta_2+\theta_3)\sin^2\theta_1\sin\theta_2 \\
 & -l_1l_2l_3\sin(\theta_2+\theta_3)\sin^2\theta_1\cos\theta_2 | \\
 & =|l_2l_3^2\cos^2(\theta_2+\theta_3)\sin\theta_2-l_2^2l_3\sin(\theta_2+\theta_3)\cos^2\theta_2 \\
 & -l_2l_3^2\cos(\theta_2+\theta_3)\sin(\theta_2+\theta_3)\cos\theta_2 \\
 & +l_2^2l_3\cos(\theta_2+\theta_3)\cos\theta_2\sin\theta_2 \\
 & +l_1l_2l_3\cos(\theta_2+\theta_3)\sin\theta_2-l_1l_2l_3\sin(\theta_2+\theta_3)\cos\theta_2 |\\
 & =
\begin{vmatrix}
-l_2l_3
\begin{pmatrix}
l_1\sin\theta_3-l_3\sin\theta_2+l_2\cos\theta_2\sin\theta_3+ \\
l_3\cos^2\theta_3\sin\theta_2+l_3\cos\theta_2\cos\theta_3\sin\theta_3
\end{pmatrix}
\end{vmatrix}
\end{aligned}
\end{equation}
From the above equation, it is clear that the manipulability \( w \) depends only on the lengths of the leg segments \( l_1 \), \( l_2 \), and \( l_3 \), and the joint angles \( \theta_2 \) and \( \theta_3 \). Therefore, subsequent optimizations will be based on the manipulability \( w \) to determine the optimal size ratios of the leg segments.
\subsection{Optimal Leg Pose and Dimensional Proportions}
Assume the total length of the leg, consisting of the coxa, femur, and tibia, is a fixed value of 1000 mm. When the total length of the leg remains constant, the segment proportions are optimized based on the manipulability \( w \). Since the swing phase leg reaches a singularity position when the knee joint angle \( \theta_3 \) is 0, the manipulability \( w \) is 0 at this point. Therefore, the range of \( \theta_3 \) is set to \([-3\pi/4, -\pi/12]\) to avoid singularities, while the range of \( \theta_2 \) remains unchanged. First, the relationship between the proportion of \( l_1 \) and the manipulability \( w \) is investigated. Keeping the ratio of \( l_2 \) to \( l_3 \) at 1:1, different proportions of \( l_1 \) relative to \( l \) are considered. Substituting the values from Table 5 into Equation (24), the corresponding manipulability surface as a function of joint angle variations is plotted using MATLAB, as shown in Fig.10.
\begin{figure}[htb]
      \centering
      \includegraphics[width=0.7\columnwidth]{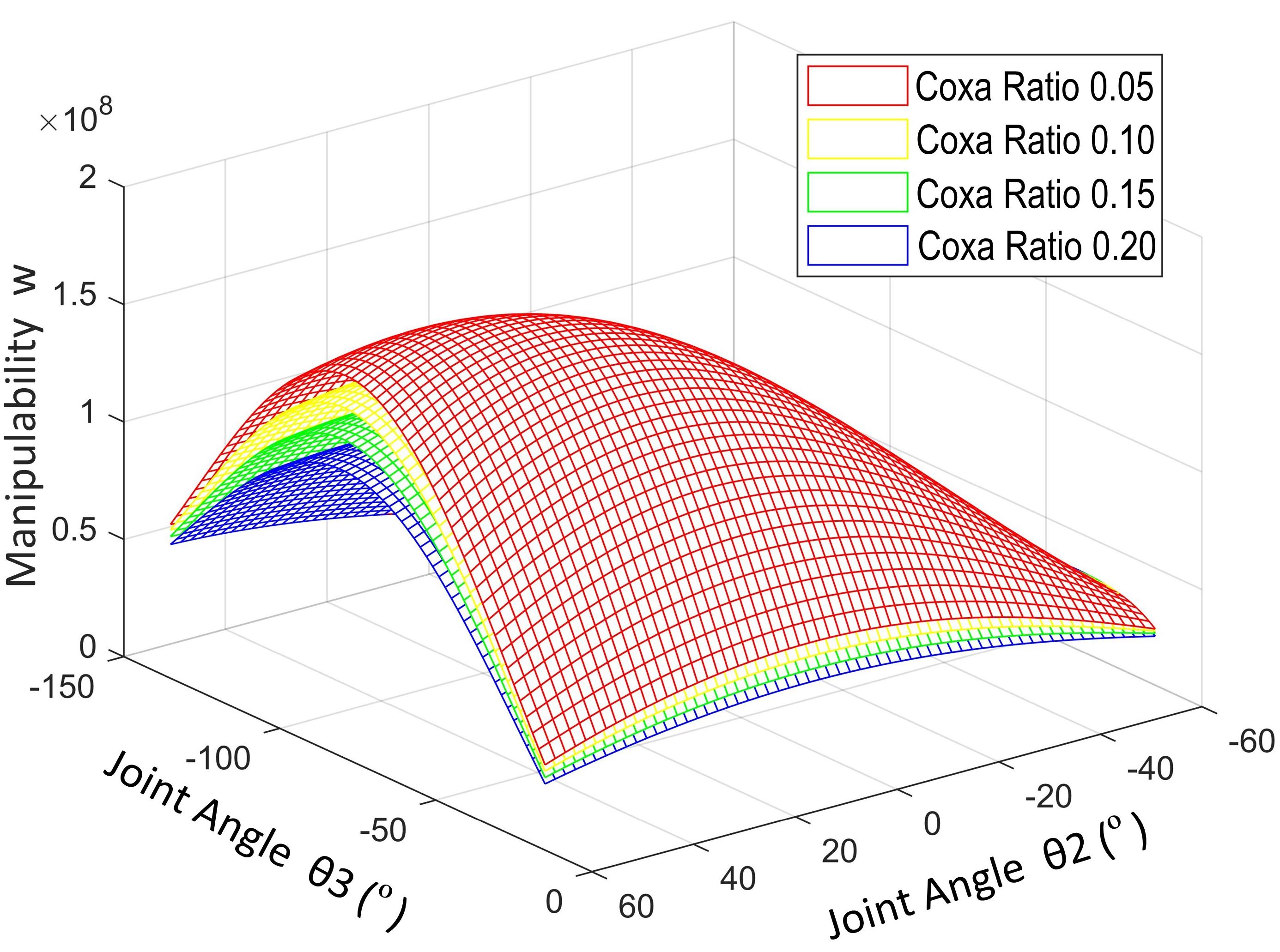}
      \caption{Manipulability Variation Surface}
      \label{Fig.10}
   \end{figure}
It can be visually observed from the figure that the manipulability \( w \) increases as the ratio of the coxa length \( l_1 \) decreases, i.e., it increases as the total length of the femur \( l_2 \) and tibia \( l_3 \) increases.

Setting the ratio of the coxa length \( l_1 \) to the total length as a constant, the relationship between the manipulability \( w \) reaching its maximum value and the dimension ratios of the femur length \( l_2 \) and tibia length \( l_3 \) is investigated. For the problem of finding the optimal value of \( w \), it can be converted into a constrained nonlinear optimization problem with multiple variables, where:

\begin{equation}
\label{(25)}
\begin{split}
w &= \lvert -l_2 l_3 (l_1 \sin\theta_3 - l_3 \sin\theta_2 + l_2 \cos\theta_2 \sin\theta_3 \\
  &\quad + l_3 \cos^2\theta_3 \sin\theta_2 + l_3 \cos\theta_2 \cos\theta_3 \sin\theta_3) \rvert
\end{split}
\end{equation}

The constraints are:
\begin{equation}
\label{(26)}
s.t.
\begin{cases}
l_1\geq0,l_2\geq0,l_3\geq0 \\
l_1+l_2+l_3=1000 \\
-60^\circ\leq\theta_2\leq60^\circ \\
-135^\circ\leq\theta_3\leq-15^\circ & 
\end{cases}
\end{equation}
Substituting the values of \( l_1 \) from Table 5 into the objective function and constraint equations, use the `fmincon` function in Matlab to solve for the joint angles and segment lengths when the manipulability \( w \) reaches its maximum value, along with the corresponding manipulability value. The optimal solution results are shown in Table 6.
\begin{table}[h]
\centering
\caption{Results of Optimal Manipulability Values}
\label{Table 6}
\begin{tabular}{cccccc}
\hline
$l_1$ (mm) & $l_2$ (mm) & $l_3$ (mm) & $\theta_2$ ($^\circ$) & $\theta_3$ ($^\circ$) & $w$ \\
\hline
50 & 475 & 475 & 35.6827 & -71.3655 & 1.7576e+08 \\
100 & 450 & 450 & 36.1108 & -72.2215 & 1.5949e+08 \\
150 & 425 & 425 & 36.5485 & -73.0969 & 1.4393e+08 \\
200 & 400 & 400 & 36.9956 & -73.9913 & 1.2903e+08 \\
\hline
\end{tabular}
\end{table}
It can be observed that under the given ratio of the coxa length \( l_1 \), the optimal configuration for the swing phase leg when the manipulability \( w \) is maximized is as follows: the hip joint angle \( \theta_2 \) is approximately 36° to 37°, and it slightly increases as the ratio of the coxa length \( l_1 \) increases; the knee joint angle \( \theta_3 \) is approximately -73° to -71°, and it slightly decreases as the ratio of the coxa length \( l_1 \) increases. For different ratios of the coxa length \( l_1 \), when the femur length \( l_2 \) to tibia length \( l_3 \) ratio is 1:1, the manipulability \( w \) of the swing phase leg in its optimal configuration reaches its maximum value.

\subsection{Average Manipulability Index and Optimization Results}
Optimization based on the maximum manipulability at the optimal configuration provides insights into the segment proportions but does not fully describe the relationship between manipulability and segment size ratios when the swing phase leg moves throughout its entire workspace. To systematically study the manipulability of the robot, combining the workspace of the leg, the concept of average manipulability \( w_a \) is introduced. This is achieved by uniformly sampling the joint angles of the swing phase leg across \( n \) characteristic values and calculating the average of the total manipulability. The expression for the average manipulability \( w_a \) is as follows:
\begin{equation}
\label{(27)}
w_A = \frac{\sum_{i=1}^m \sum_{j=1}^n w\bigl(\theta_{2i}, \theta_{3j}\bigr)}{mn}
\end{equation}
Where \( m \) and \( n \) represent the number of uniformly distributed characteristic values of the joint angles \( \theta_2 \) and \( \theta_3 \) within their respective ranges; \( \theta_{2i} \) denotes the \( i \)-th characteristic value of \( \theta_2 \), and \( \theta_{3j} \) denotes the \( j \)-th characteristic value of \( \theta_3 \).

Based on the specified ranges for the joint angles \( \theta_2 \) and \( \theta_3 \), with a characteristic value selected every 1°, \( m = 121 \) and \( n = 121 \). Substituting these into the calculations and plotting the changes in \( w_a \) with respect to the ratio of the coxa length \( l_1 \) to the tibia length \( l_3 \), as shown in Fig.11.
\begin{figure}[htb]
      \centering
      \includegraphics[width=0.7\columnwidth]{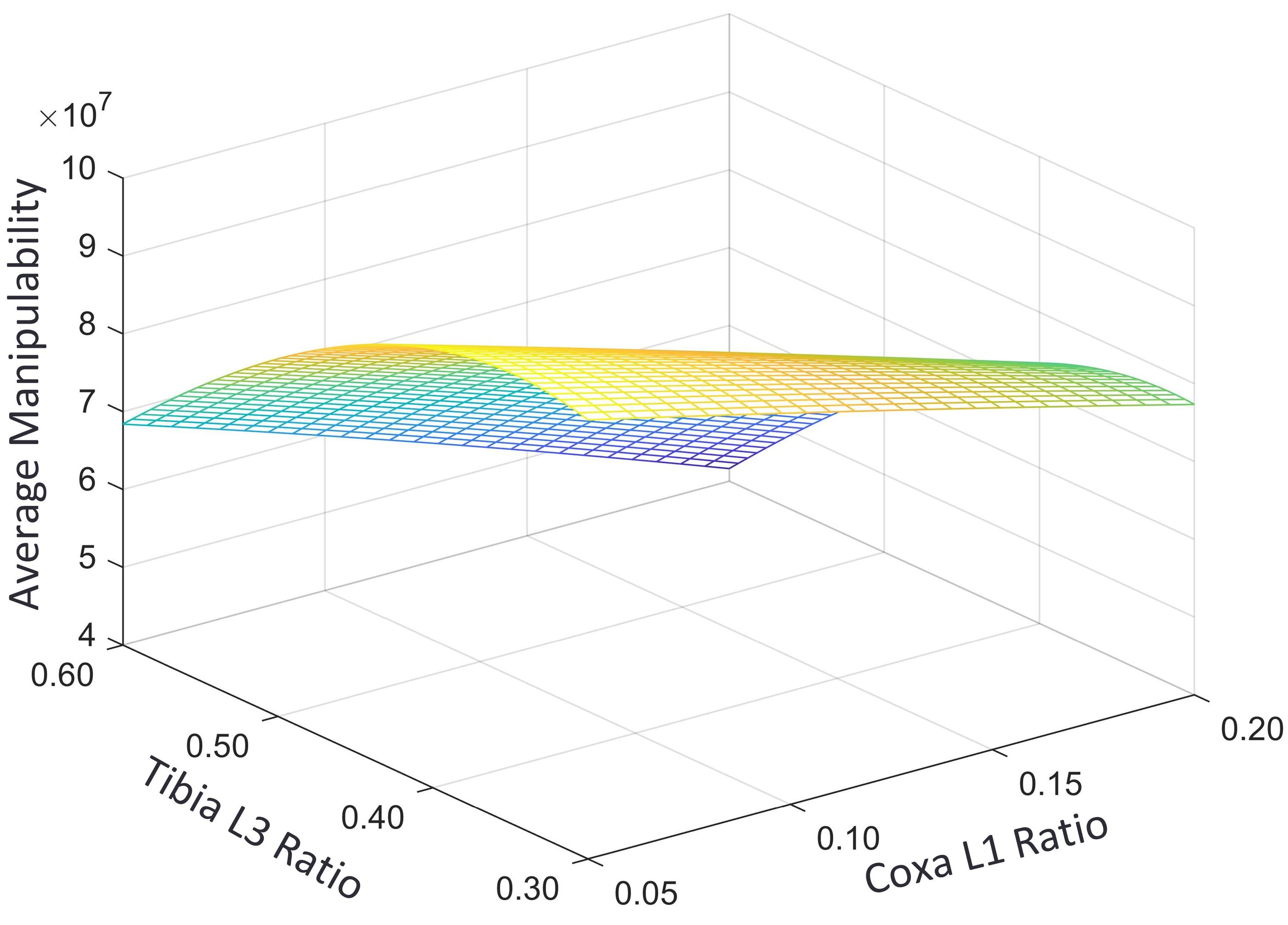}
      \caption{Surface of Average Manipulability Variation}
      \label{Fig.11}
   \end{figure}
It can be observed that the average manipulability \( w_a \) decreases as the ratio of the coxa length \( l_1 \) increases. Furthermore, when the ratio of the tibia length \( l_3 \) is approximately 0.45, the average manipulability \( w_a \) reaches its maximum value.
\section{Dimension Optimization Based on Body Flexibility}
\subsection{Flexibility of the Robot Body}
The previous discussion primarily focused on the kinematic performance of the swing phase leg. However, the structural parameters of the stance phase leg can also affect the kinematic performance of the robot's body frame. The position of the robot's body in a multi-legged robot can be represented by six parameters: three translations \( x \), \( y \), \( z \) along the XYZ axes relative to the ground coordinate system, and three rotations \( \alpha \), \( \beta \), \( \gamma \) about the XYZ axes. When the body position of a hexapod robot is adjusted, these six parameters will vary within certain ranges. Let \( S_x \), \( S_y \), \( S_z \), \( \phi_x \), \( \phi_y \), \( \phi_z \) represent the intervals of variation for the translational and rotational displacements along the XYZ axes, respectively.
\begin{equation}
\label{(28)}
\begin{cases}
\quad S_x = X_{\max} - X_{\min}, \quad S_y = Y_{\max} - Y_{\min}, \quad S_z = Z_{\max} - Z_{\min},\\
\quad \phi_x = \alpha_{\max} - \alpha_{\min}, \quad \phi_y = \beta_{\max} - \beta_{\min}, \quad \phi_z = \gamma_{\max} - \gamma_{\min}.
\end{cases}
\end{equation}
Let the flexibility be denoted as \( FB \), and its expression is:
\begin{equation}
\label{(29)}
FB=\frac{1}{6}{\left[\frac{1}{2L}{\left(S_x+S_y+S_z\right)}+\frac{1}{180}{\left(\phi_x+\phi_y+\phi_z\right)}\right]}
\end{equation}
In the equation, \( FB \) is a dimensionless constant ranging between 1 and 0, reflecting the overall flexibility of the hexapod robot; \( L \) represents the leg length.

Based on the forward and inverse kinematic analysis of the support phase legs, it is known that when the number of support phase legs is 3, there exists a corresponding relationship between the foot-end position coordinates, joint angles, and the pose parameters of the body frame. Similarly, the kinematic relationships for when the number of support phase legs is 6 can be derived as follows:
\begin{equation}
\label{(30)}
\begin{split}
& 
\begin{bmatrix}
p_{x1} & p_{x2} & p_{x3} & p_{x4} & p_{x5} & p_{x6} \\
p_{y1} & p_{y2} & p_{y3} & p_{y4} & p_{y5} & p_{y6} \\
p_{z1} & p_{z2} & p_{z3} & p_{z4} & p_{z5} & p_{z6} \\
1 & 1 & 1 & 1 & 1 & 1
\end{bmatrix} \\
& =T_C
\begin{bmatrix}
u_1 & u_2 & u_3 & u_4 & u_5 & u_6 \\
v_1 & v_2 & v_3 & v_4 & v_5 & v_6 \\
w_1 & w_2 & w_3 & w_4 & w_5 & w_6 \\
1 & 1 & 1 & 1 & 1 & 1
\end{bmatrix}
\end{split}
\end{equation}
For a hexapod robot with all six feet in contact with the ground, the foot-end position coordinates and joint angles are known quantities. Kinematic analysis using the above equation theoretically allows for the calculation of parameters required for manipulability. However, the robot is equivalent to a redundant parallel mechanism at this point, where the posture of each supporting leg may differ. The extreme positions of the body can only be reached when some joints of certain legs reach their limit values. Consequently, the theoretical analytical approach involves substantial computational complexity and is prone to deviations from actual values. To address the challenge of analyzing the kinematic relationships in such complex mechanisms, a virtual prototyping simulation method can be employed to study the relationship between the body's flexibility and leg dimensions.
\subsection{Virtual Prototyping and Model Development}
This section employs virtual prototyping technology to analyze the body flexibility of the construction robot. By parameterizing the robot's leg structure, the lengths of each leg segment are treated as variables. Through simulation experiments, the dimensions of the leg segments are optimized.

(1)Establishment of the Robot's Parameterized Model

A parameterized model of the hexapod robot is established using ADAMS. Models for the body, ground, and legs are created, with the coxa length, femur length, and tibia length set as parameterized variables. Based on the leg kinematic model, single-degree-of-freedom revolute joints are added between the body and the coxa, as well as between each leg segment. A three-degree-of-freedom spherical joint is added between the tibia end (foot-end) and the ground. To prevent misalignment of the kinematic joints during parameter variation, associations are established between the links of each leg model and the I and J marker points of the joints. The simplified parameterized model of the hexapod robot is shown in Fig.12.
\begin{figure}[htb]
      \centering
      \includegraphics[width=0.4\columnwidth]{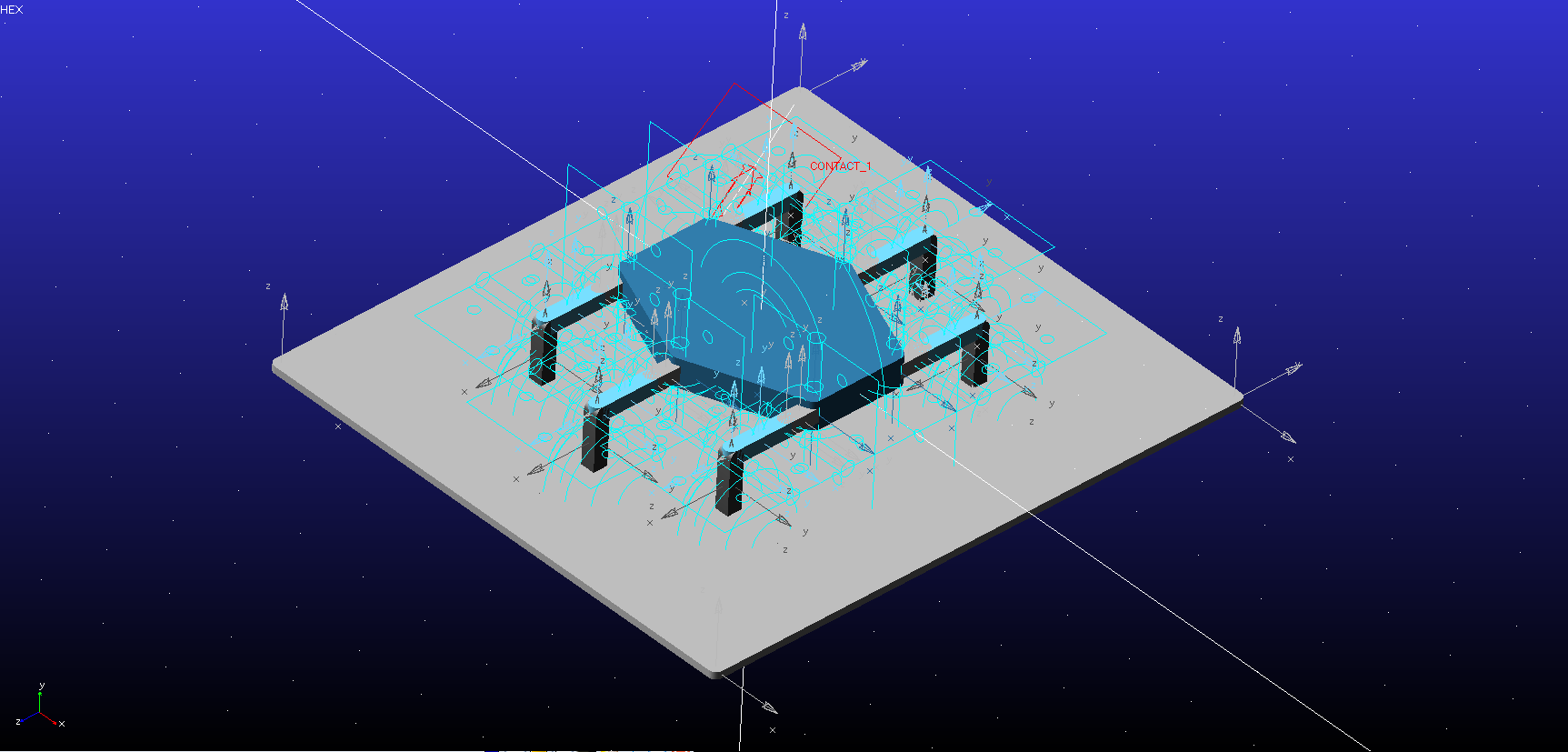}
      \caption{Parameterized Model of the Hexapod Robot}
      \label{Fig.12}
   \end{figure}

(2) Setting Joint Angle Ranges and Collision Detection

To record the parameters required for calculating the body flexibility, the geometric center of the body is set as the measurement point. Translations along the positive and negative directions of the XYZ axes and rotations about the positive and negative directions of the XYZ axes are defined, totaling 12 measurement values [30]. Each joint is equipped with a joint measurement and two sensors. When the joint angle measurement reaches the predefined limit value, the sensor triggers to stop the body movement, thereby limiting the joint range. Initially, the range for the root joint angle $ \theta_1 $ is set to $[- \pi / 4, \pi / 4]$, the hip joint angle $ \theta_2 $ to $[- \pi / 6, \pi / 6]$, and the knee joint angle $ \theta_3 $ to $[- 3\pi / 4, - \pi / 12]$. Finally, contact is added between the ground and the robot body. When a collision occurs, the body's movement also stops.

\begin{figure}[htb]
      \centering
      \includegraphics[width=\columnwidth]{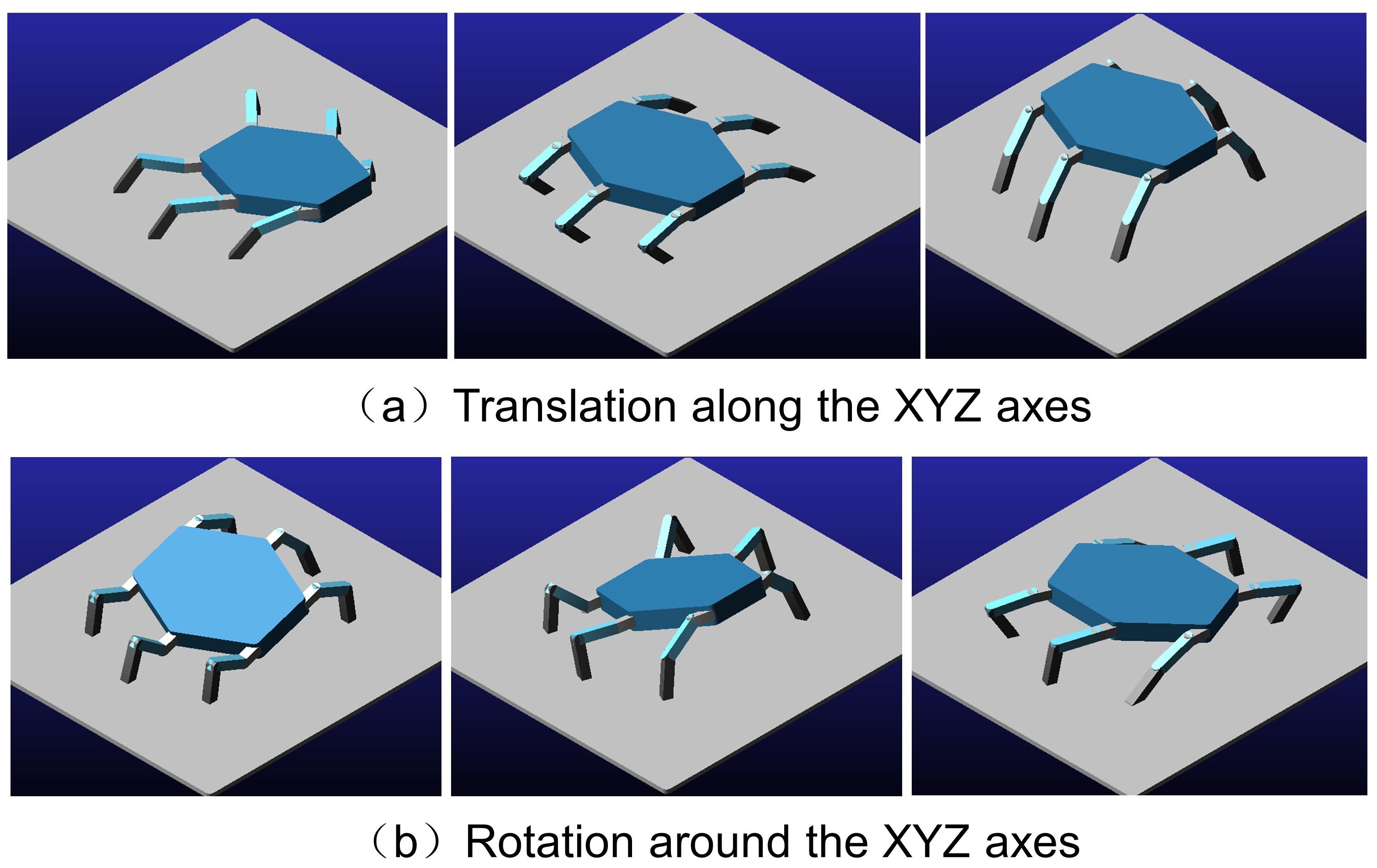}
      \caption{Schematic Diagram of the Robot Body in Extreme Poses}
      \label{Fig.13}
   \end{figure}

\begin{figure}[htbp]
    \centering
    \begin{subfigure}[t]{0.48\textwidth}
        \centering
        \includegraphics[width=\linewidth]{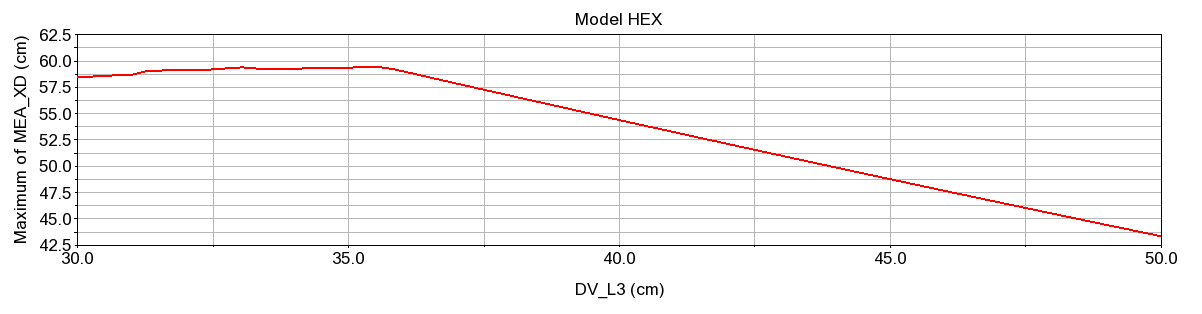}
        \caption{Positive X-axis Translational Displacement}
    \end{subfigure}
    \hspace{0.02\textwidth}
    \begin{subfigure}[t]{0.48\textwidth}
        \centering
        \includegraphics[width=\linewidth]{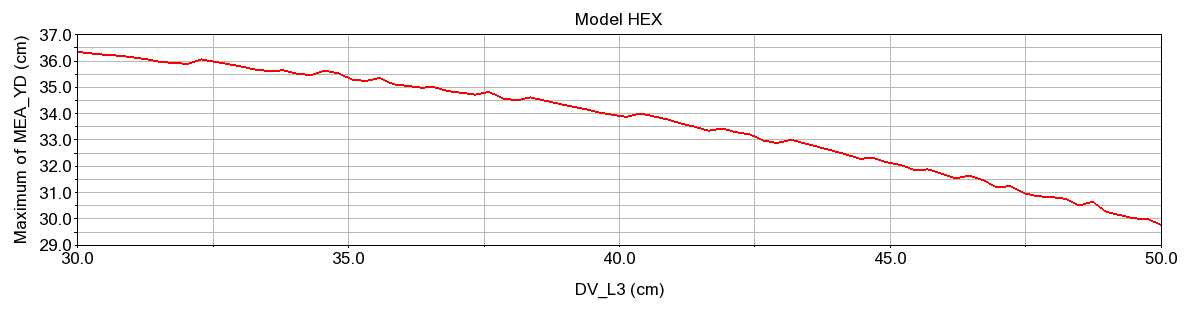}
        \caption{Positive Y-axis Translational Displacement}
    \end{subfigure}

    \vspace{0.3em}
    \begin{subfigure}[t]{0.48\textwidth}
        \centering
        \includegraphics[width=\linewidth]{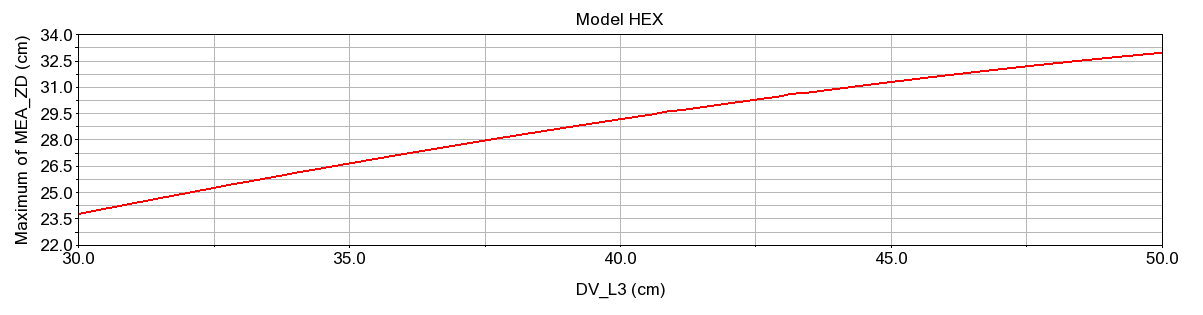}
        \caption{Positive Z-axis Translational Displacement}
    \end{subfigure}
    \hspace{0.02\textwidth}
    \begin{subfigure}[t]{0.48\textwidth}
        \centering
        \includegraphics[width=\linewidth]{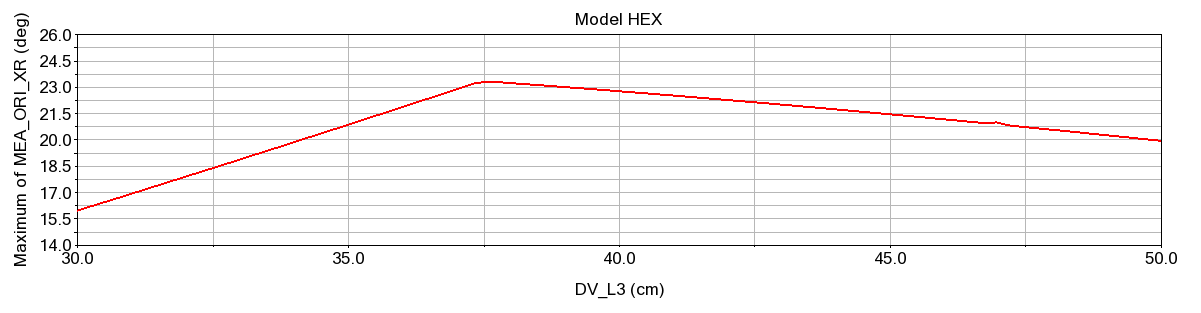}
        \caption{Rotation Around X-axis}
    \end{subfigure}

    \vspace{0.3em}
    \begin{subfigure}[t]{0.48\textwidth}
        \centering
        \includegraphics[width=\linewidth]{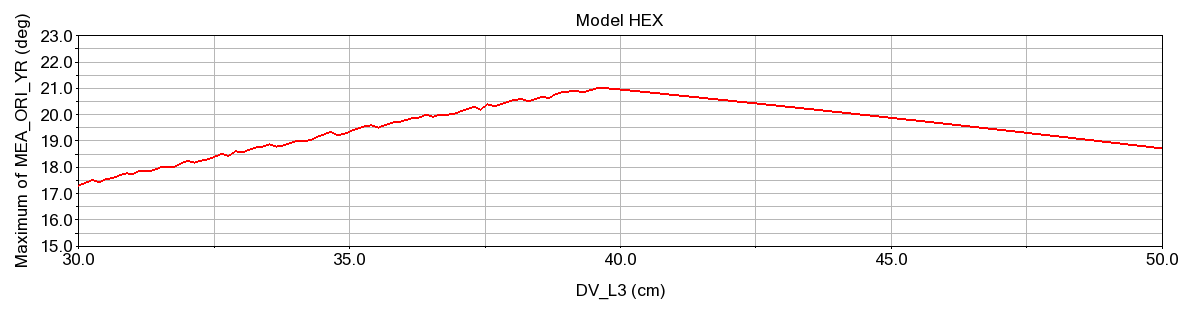}
        \caption{Rotation Around Y-axis}
    \end{subfigure}
    \hspace{0.02\textwidth}
    \begin{subfigure}[t]{0.48\textwidth}
        \centering
        \includegraphics[width=\linewidth]{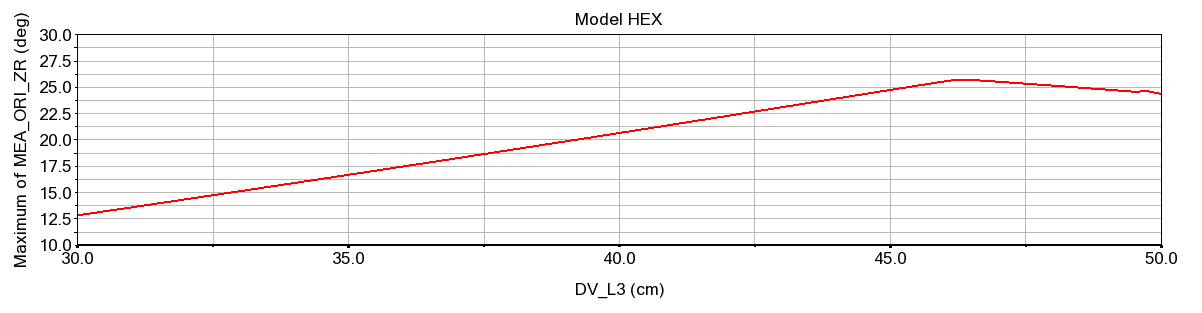}
        \caption{Rotation Around Z-axis}
    \end{subfigure}

    \caption{Curves of Translational and Rotational Displacement Variations Along Positive Axes}
    \label{Fig.14}
\end{figure}

(3) Setting Actuation and Motion Simulation

The Motion command is used to apply actuation to the robot, enabling translational and rotational movements of the body along the positive and negative directions of the XYZ axes until the body reaches its limit poses. Fig.13 illustrates the six limit poses corresponding to the body's translational and rotational movements along the positive directions of the XYZ axes.

Using the ADAMS Design Study module, the tibia length (DV-L3) is set as a variable. Similar to the previously described research method, the total leg length is fixed at 1000 mm, and the coxa length (DV-L1) is assigned different values for each experiment. The femur length (DV-L2) varies with the tibia length. The 12 parameters of the body's limit poses are recorded for each simulation, and each experiment involves 80 simulation calculations.

\subsection{Organization and Analysis of Simulation Experiment Results}
After iterative calculations in ADAMS, the software outputs the variation data of the extreme poses of each measurement with respect to the number of iterations. When the coxa length ratio is set to 0.05, the curves of translational and rotational displacements along the positive X, Y, and Z axes against tibia length are presented in Fig.14.

The coxa length ratios are set to 0.05, 0.10, 0.15, and 0.20, and simulation experiments are conducted accordingly. The 48 sets of data calculated by ADAMS are exported and organized. The curves showing the variation of body flexibility \( FB \)  with coxa length ratios and tibia length \( l_3 \) are presented in Fig.15.
\begin{figure}[htb]
      \centering
      \includegraphics[width=0.7\columnwidth]{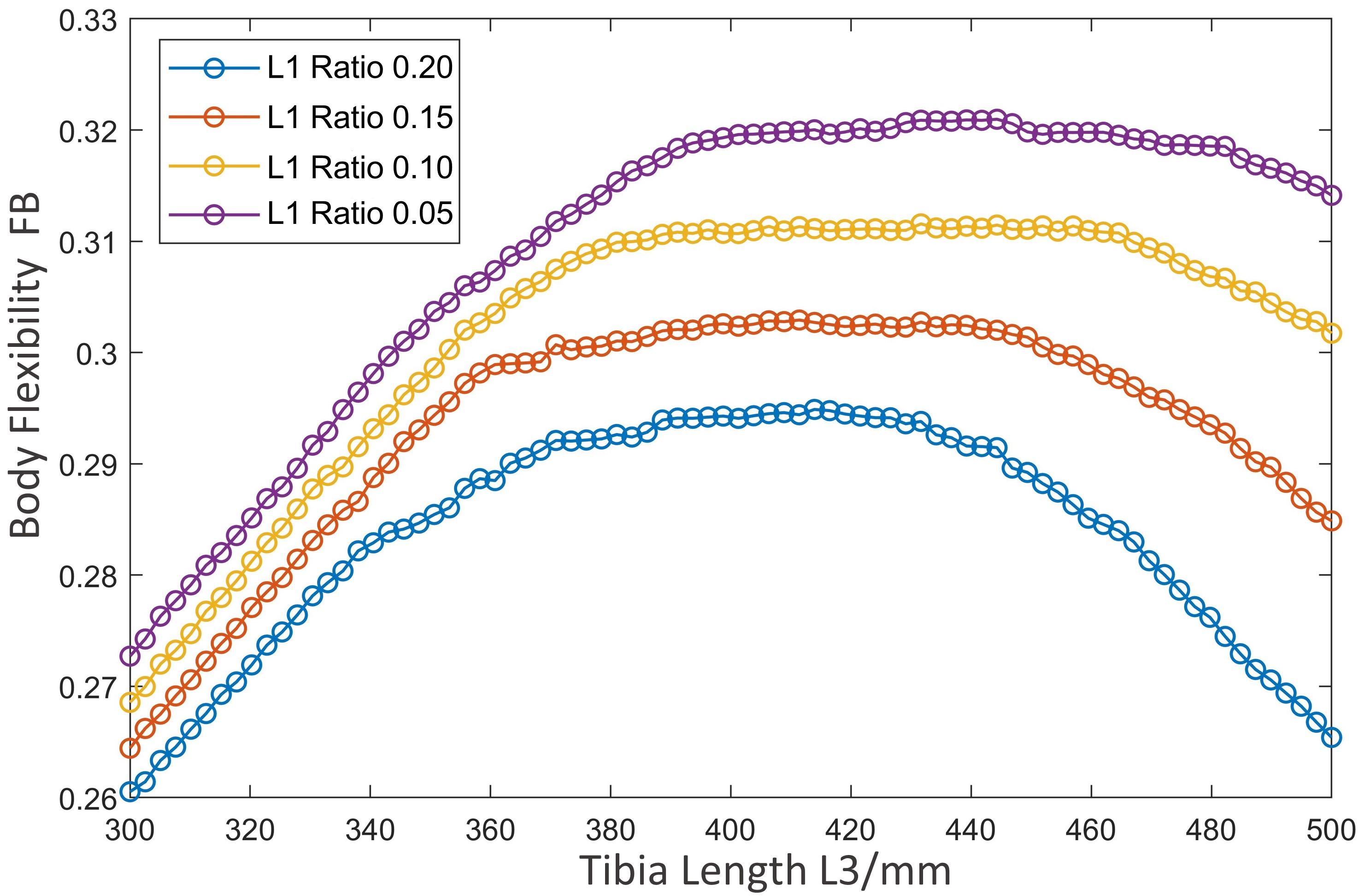}
      \caption{Curve of Body Flexibility Variation}
      \label{Fig.15}
   \end{figure}
It can be observed that as the coxa length ratio \( l_1 \) decreases, the body flexibility increases. When the coxa length is constant, the body flexibility reaches its maximum value when the tibia length ratio \( l_3 \) is approximately between 0.4 and 0.45.

Combining the previous results based on workspace area and average manipulability for size optimization, it is evident that when the total leg length \( l \) of the robot is fixed, the coxa length ratio \( l_1 \) should be as small as possible. Specifically, when the coxa length ratio \( l_1 \) is between 0.1 and 0.2, setting the tibia length ratio \( l_3 \) between 0.4 and 0.45 yields better overall kinematic performance for both the robot's legs and the entire system.

\section{Conclusions}

Construction robots represent a new direction in the future development of robotics. Focusing on leg optimization, a series of innovative design and optimization methods are proposed. Inspired by multi-legged insects in nature, the leg architecture and single-leg structure of construction robots are designed. Based on an in-depth analysis of the kinematic characteristics of the swing phase leg, this paper introduces the concept of an "improved workspace" and uses graphical methods to determine the optimal length ratios of each leg segment. Additionally, the metric of "average manipulability" is introduced based on the velocity Jacobian matrix, and numerical solutions are employed to derive the leg segment ratios that achieve optimal manipulability. 

For the stance phase legs, ADAMS virtual prototyping is utilized for simulation experiments to explore the relationship between the body flexibility of the construction robot and the proportions of its leg segments. This further validates the guiding role of kinematic metrics in practical design. By integrating the optimization results of various metrics, this study achieves, for the first time, a multidimensional quantitative evaluation of the kinematic performance of hexapod robot legs for complex construction scenarios. This provides new insights into overcoming the limitations of traditional wheeled or tracked robots in unstructured terrains and dynamic obstacle environments. It also offers robust support for the multi-objective optimization of key performance indicators in construction robots and points the way for future research on robot design and control in more complex environments.

\bigskip 
\noindent \textbf{Acknowledgements} \\
This work was supported by the National Key R\&D Program of China under Grant No.2023YFB4705002; National Natural Science Foundation of China under Grant No.U20A20283; Guangdong Provincial Key Laboratory of Construction Robotics and Intelligent Construction under Grant No.2022KSYS013; CAS Science and Technology Service Network Plan (STS) - Dongguan Special Project under Grant No.20211600200062; and the Science and Technology Cooperation Project of Chinese Academy of Sciences in Hubei Province Construction 2023.

\end{document}